\documentclass[10pt,journal,compsoc]{IEEEtran}
\usepackage{times}
\usepackage{epsfig}
\usepackage{graphicx}
\usepackage{amsmath}
\usepackage{amssymb}
\usepackage{algorithm}
\usepackage{algpseudocode}
\DeclareMathOperator*{\argmin}{argmin}   
\usepackage{mwe}
\usepackage{booktabs} 

\usepackage[hyphens]{url}
\usepackage[flushleft]{threeparttable}
\usepackage{tablefootnote}
\usepackage{mathtools}
\DeclarePairedDelimiter \floor{\lfloor}{\rfloor}

\usepackage[pagebackref=true,breaklinks=true,letterpaper=true,colorlinks,bookmarks=false]{hyperref}

\ifCLASSOPTIONcompsoc
  \usepackage[nocompress]{cite}
\else
  \usepackage{cite}
\fi

\ifCLASSINFOpdf
\else
\fi

\ifCLASSOPTIONcompsoc
  \usepackage[caption=false,font=footnotesize,labelfont=sf,textfont=sf]{subfig}
\else
  \usepackage[caption=false,font=footnotesize]{subfig}
\fi

\begin{document}
%

\title{\huge Learning a Fixed-Length Fingerprint Representation}
%
%
%
%

\author{Joshua~J.~Engelsma,
        Kai~Cao,
        and~Anil~K.~Jain,~\IEEEmembership{Life~Fellow,~IEEE}
\IEEEcompsocitemizethanks{
\IEEEcompsocthanksitem J. J. Engelsma and A. K. Jain are with the Department of Computer Science and Engineering, Michigan State University, East Lansing, MI, 48824\protect\\
E-mail: engelsm7@msu.edu, jain@cse.msu.edu

\IEEEcompsocthanksitem K. Cao is a Senior Biometrics Researcher at Goodix, San Diego, CA\protect\\
E-mail: caokai0505@gmail.com
}

}

%
%

\markboth{IEEE Transactions on Pattern Analysis and Machine Intelligence}%
{Engelsma \MakeLowercase{\textit{et al.}}: IEEE Transactions on Pattern Analysis and Machine Intelligence}
%




\IEEEtitleabstractindextext{%
\begin{abstract}
We present DeepPrint, a deep network, which learns to extract fixed-length fingerprint representations of only 200 bytes. DeepPrint incorporates fingerprint domain knowledge, including alignment and minutiae detection, into the deep network architecture to maximize the discriminative power of its representation. The compact, DeepPrint representation has several advantages over the prevailing variable length minutiae representation which (i) requires computationally expensive graph matching techniques, (ii) is difficult to secure using strong encryption schemes (e.g. homomorphic encryption), and (iii) has low discriminative power in poor quality fingerprints where minutiae extraction is unreliable. We benchmark DeepPrint against two top performing COTS SDKs (Verifinger and Innovatrics) from the NIST and FVC evaluations. Coupled with a re-ranking scheme, the DeepPrint rank-1 search accuracy on the NIST SD4 dataset against a gallery of 1.1 million fingerprints is comparable to the top COTS matcher, but it is significantly faster (\textbf{DeepPrint:} 98.80\% in 0.3 seconds vs. \textbf{COTS A:} 98.85\% in 27 seconds). To the best of our knowledge, the DeepPrint representation is the most compact and discriminative fixed-length fingerprint representation reported in the academic literature.
\end{abstract}

}

\maketitle

\IEEEdisplaynontitleabstractindextext

%
\IEEEpeerreviewmaketitle

\IEEEraisesectionheading{\section{Introduction}\label{sec:introduction}}

\begin{figure}[t]
  \centering
  \subfloat[Level-1 features]{\includegraphics[scale=0.52]{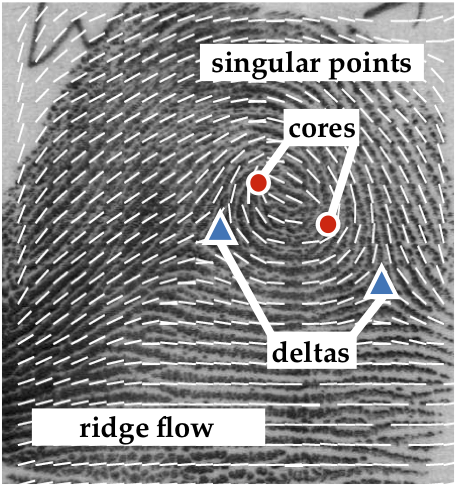}\label{fig:f1}}
  \hfill
  \subfloat[Level-2 features]{\includegraphics[scale=0.52]{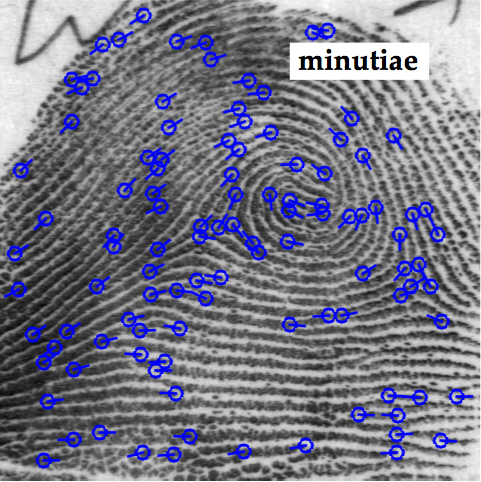}\label{fig:f2}}
  \caption{The most popular fingerprint representation consists of (a) global level-1 features (ridge flow, core, and delta) and (b) local level-2 features, called minutiae points, together with their descriptors (e.g., texture in local minutiae neighborhoods). The fingerprint image illustrated here is a rolled impression from the NIST SD4 database~\cite{sd4}. The number of minutiae in NIST4 rolled fingerprint images range all the way from 12 to 196.}
  \label{fig:intro}
  \vspace{-1.0em}
\end{figure}

%
%
%
%
\IEEEPARstart{O}{ver} 100 years ago, the pioneering giant of modern day fingerprint recognition, Sir Francis Galton, astutely commented on fingerprints in his 1892 book titled ``Finger Prints": 

\begin{quote}
\textit{``They have the unique merit of retaining all their peculiarities unchanged throughout life, and afford in consequence an incomparably surer criterion of identity than any other bodily feature."}~\cite{galton}
\end{quote} Galton went on to describe fingerprint \textit{minutiae}, the small details woven throughout the papillary ridges on each of our fingers, which Galton believed provided uniqueness and permanence properties for accurately identifying individuals. Over the 100 years since Galton's ground breaking scientific observations, fingerprint recognition systems have become ubiquitous and can be found in a plethora of different domains~\cite{handbook} such as forensics~\cite{ngi}, healthcare, mobile device security~\cite{touchid},  mobile payments~\cite{touchid}, border crossing~\cite{obim}, and national ID~\cite{india1}. To date, virtually all of these systems continue to rely upon the location and orientation of minutiae within fingerprint images for recognition (Fig.~\ref{fig:intro}).

\newcommand{\specialcell}[2][c]{%
  \begin{tabular}[#1]{@{}c@{}}#2\end{tabular}}
  \newcommand{\tabitem}{~~\llap{\textbullet}~~}

Although automated fingerprint recognition systems based on minutiae representations (\textit{i.e.} handcrafted features) have seen tremendous success over the years, they have several limitations.  

\begin{figure}[t]
\begin{center}
\includegraphics[scale=0.5]{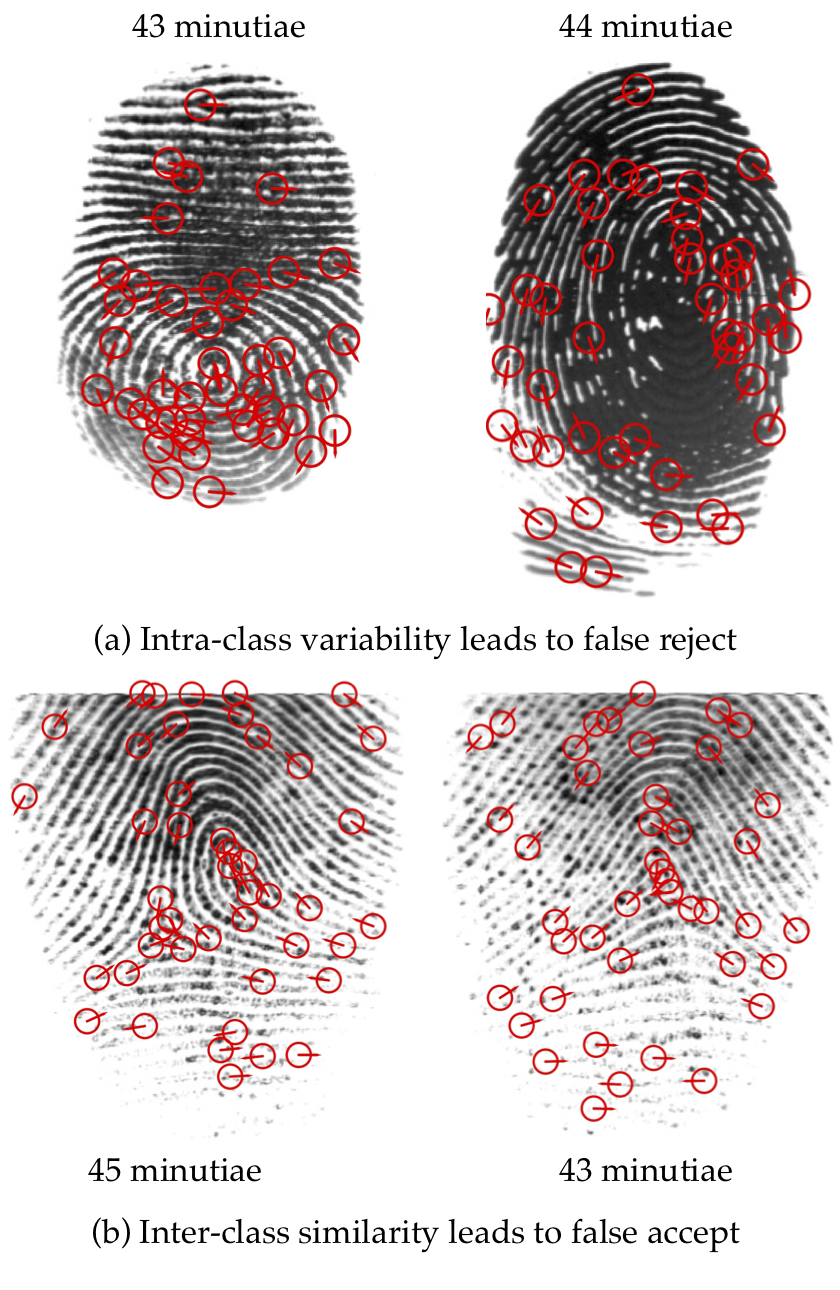}
\caption{Failures of the COTS A minutiae-based matcher (minutiae annotated with COTS A). The genuine pair (two impressions from the same finger) in (a) was falsely rejected at 0.1\% FAR (score of 9) due to heavy non-linear distortion and moist fingers. The imposter pair (impressions from two different fingers) in (b) was falsely accepted at 0.1\% FAR (score of 38) due to the similar minutiae distribution in these two fingerprint images (the score threshold for COTS A @ FAR = 0.1\% is 34). In contrast, DeepPrint is able to correctly match the genuine pair in (a) and reject the imposter pair in (b). These slap fingerprint impressions come from public domain FVC 2004 DB1 A database~\cite{fvc2004}. The number of minutiae in FVC 2004 DB1 A images range from 11 to 87.}
\label{fig:intro_fig}
\end{center}
\vspace{-1.0em}
\end{figure} 

\begin{itemize}
\item Minutiae-based representations are of variable length, since the number of extracted minutiae (Table~\ref{table:templates}) varies amongst different fingerprint images even of the same finger (Fig.~\ref{fig:intro_fig} (a)). Variations in the number of minutiae originate from a user's interaction with the fingerprint reader (placement position and applied pressure) and condition of the finger (dry, wet, cuts, bruises, etc.). This variation in the number of minutiae causes two main problems: (i) pairwise fingerprint comparison is computationally demanding and varies with number of minutiae and (ii) matching in the encrypted domain, a necessity for user privacy protection, is computationally expensive, and results in loss of accuracy~\cite{encryption}.

\item In the context of global population registration, fingerprint recognition can be viewed as a \textbf{75 billion class problem} ($\approx7.5$ billion living persons, assuming nearly all with 10 fingers) with large intra-class variability and large inter-class similarity (Fig.~\ref{fig:intro_fig}). This necessitates extremely discriminative yet compact representations that are complementary and at least as discriminative as the traditional minutiae-based representation. For example, India's civil registration system, Aadhaar, now has a database of $\approx1.3$ billion residents who are enrolled based on their 10 fingerprints, 2 irises, and face image~\cite{india1}.

\item Reliable minutiae extraction in low quality fingerprints (due to noise, distortion, finger condition) is problematic, causing false rejects in the recognition system (Fig.~\ref{fig:intro_fig} (a)). See also NIST fingerprint evaluation FpVTE 2012~\cite{nist}.
\end{itemize}

\begin{table}[t]
\small
\caption{Comparison of variable length minutiae representation with fixed-length DeepPrint representation}
 \centering
\begin{threeparttable}
\begin{tabular}{ccc}
 \toprule
 Matcher & \specialcell{(Min, Max) \\ \# of Minutiae\tnote{1}}  & \specialcell{(Min, Max) \\Template Size (kB)} \\
 \midrule
 \specialcell{COTS A} & (12, 196) & (1.5, 23.7) \\
  \midrule
 \specialcell{COTS B} & (12, 225) & (0.6, 5.3) \\
  \midrule
   \specialcell{Proposed} & N.A.\tnote{2} & 0.2\tnote{\textdagger} \\
  \bottomrule
\end{tabular}
\begin{tablenotes}
\item[1] Statistics from NIST SD4 and FVC 2004 DB1.
\item[2] Template is not explicitly comprised of minutiae.
\item[\textdagger] Template size is fixed at 200 bytes, irrespective of the number of minutiae (192 bytes for the features and 8 bytes for 2 decompression scalars).
\end{tablenotes}
\end{threeparttable}
\label{table:templates}
\vspace{-1.em}
\end{table}

\begin{figure}[t]
\begin{center}
\includegraphics[scale=0.46]{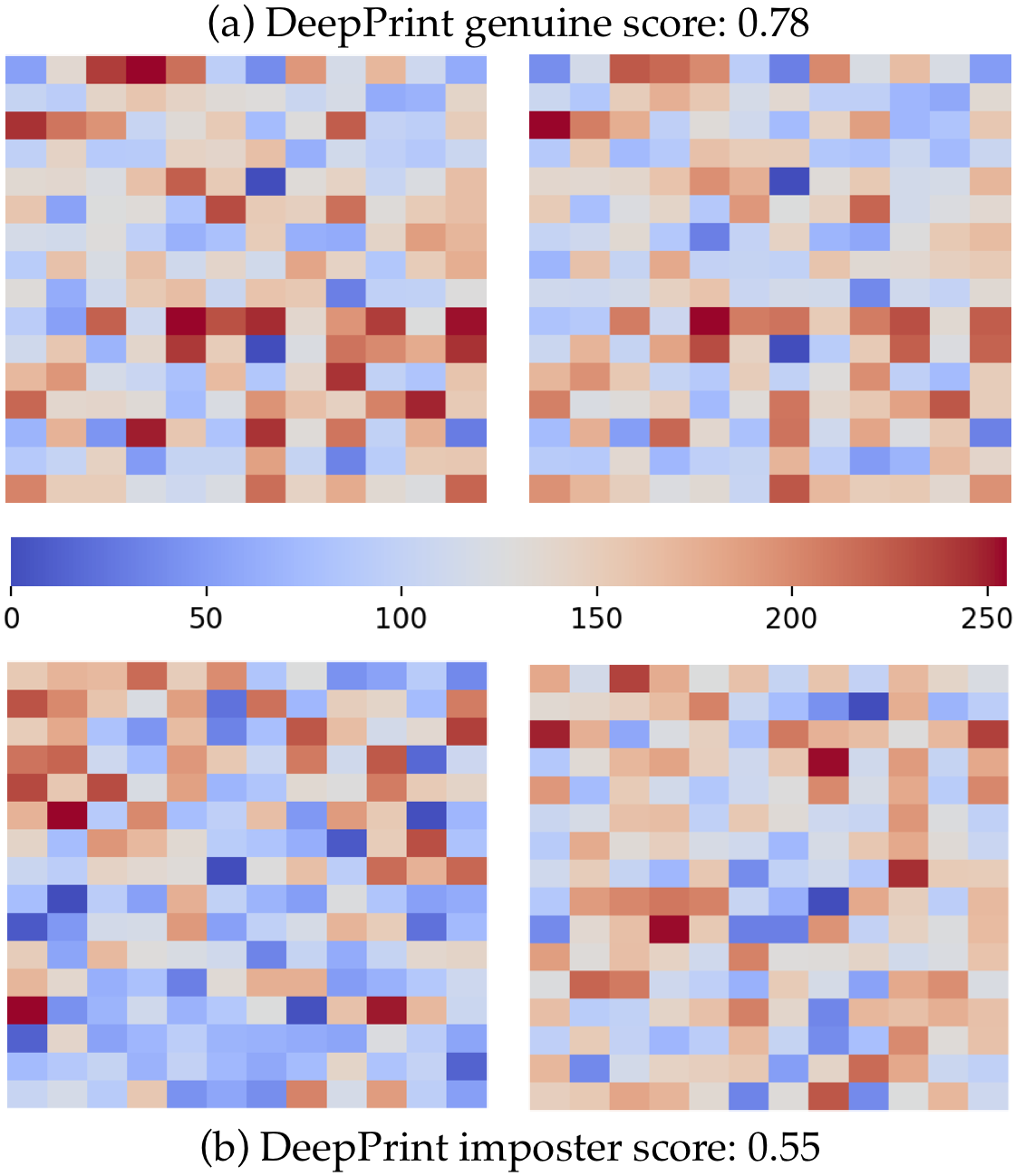}
\caption{Fixed-length, 192-dimensional fingerprint representations extracted by DeepPrint (shown as $16\times12$ feature maps) from the same four fingerprints shown in Figure~\ref{fig:intro_fig}. Unlike COTS A, we correctly
classify the pair in (a) as a genuine pair, and the pair in (b)
as an imposter pair. The score threshold of DeepPrint @
FAR = 0.1\% is 0.76}
\label{fig:feature_maps}
\end{center}
\vspace{-1.0em}
\end{figure} 

\begin{figure*}[t]
\begin{center}
\includegraphics[scale=0.5]{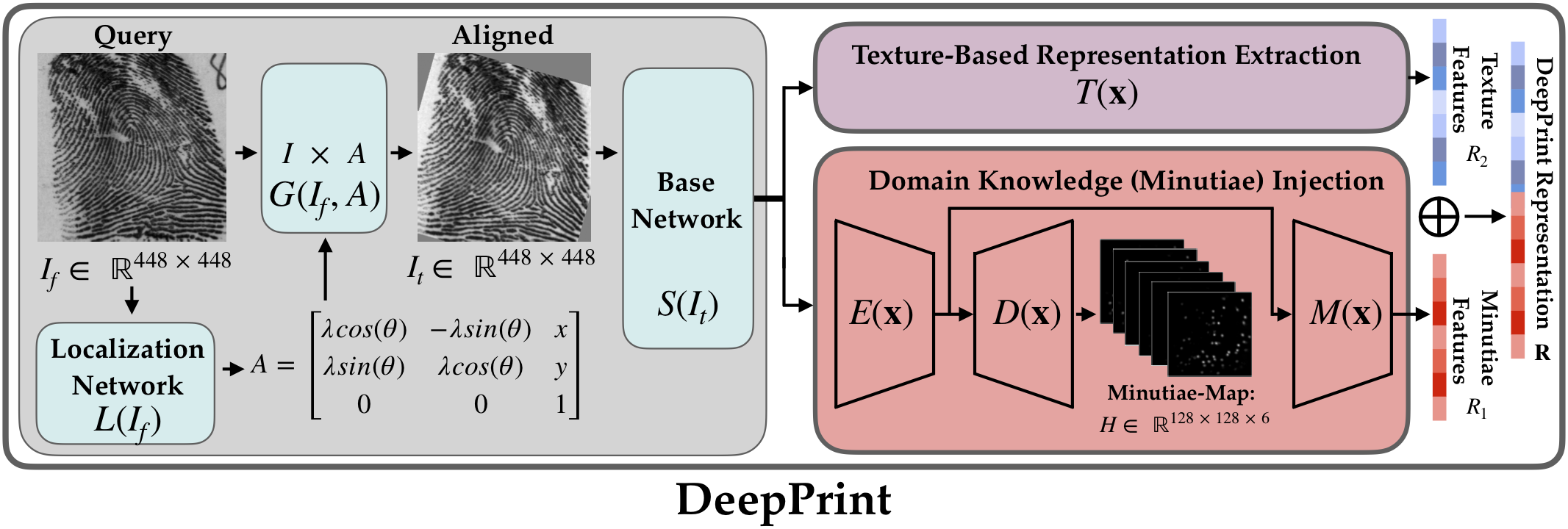}
\caption{Flow diagram of DeepPrint: (i) a query fingerprint is aligned via a Localization Network which has been trained end-to-end with the Base-Network and Feature Extraction Networks (no reference points are needed for alignment); (ii) the aligned fingerprint proceeds to the Base-Network which is followed by two branches; (iii) the first branch extracts a 96-dimensional texture-based representation; (iv) the second branch extracts a 96-dimensional minutiae-based representation, guided by a side-task of minutiae detection (via a minutiae map which does not have to be extracted during testing); (v) the texture-based representation and minutiae-based representation are concatenated into a 192-dimensional representation of 768 bytes (192 features and 4 bytes per float). The 768 byte template is compressed into a 200 byte fixed-length representation by truncating floating point value features into integer value features, and saving the scaling and shifting values (8 bytes) used to truncate from floating point values to integers. The 200 byte DeepPrint representations can be used both for authentication and large-scale fingerprint search. The minutiae-map can be used to further improve system accuracy and interpretability by re-ranking candidates retrieved by the fixed-length representation.}
\label{fig:schematic}
\end{center}
\end{figure*} 

To overcome the limitations of minutiae-based matchers, we present a reformulation of the fingerprint recognition problem. In particular, rather than extracting varying length minutiae-sets for matching (\textit{i.e.} handcrafted features), we design a deep network embedded with fingerprint domain knowledge, called \textbf{DeepPrint}, to \textit{learn} a fixed-length representation of 200 bytes which discriminates between fingerprint images from different fingers (Fig.~\ref{fig:schematic}). Our work follows the trajectory of state-of-the-art automated face recognition systems which have almost entirely abandoned traditional handcrafted features in favor of deep features extracted by deep networks with remarkable success~\cite{face1, face3, face5}. However, unlike deep network based face recognition systems, we do not completely abandon handcrafted features. Instead, we aim to integrate handcrafted fingerprint features (minutiae~\footnote{Note that we do not \textit{require} explicitly storing minutiae in our final template. Rather, we aim to guide DeepPrint to extract features related to minutiae during training of the network.}) into the deep network architecture to exploit the benefits of both deep networks and traditional, domain knowledge inspired features.    

While prevailing minutiae-matchers require expensive graph matching algorithms for fingerprint comparison, the 200 byte representations extracted by DeepPrint can be compared using simple distance metrics such as the cosine similarity, requiring only $d$ multiplications and $d - 1$ additions, where $d$ is the dimensionality of the representation (for DeepPrint, $d=192$)\footnote{The DeepPrint representation is originally 768 bytes (192 features and 4 bytes per float value). We compress the 768 bytes to 200 by scaling the floats to integer values between [0,255] and saving the two compression parameters with the features. This loss in precision (which saves significant disk storage space) very minimally effects matching accuracy.}. Another significant advantage of this fixed-length representation is that it can be matched in the encrypted domain using fully homomorphic encryption~\cite{homomorphic, homo1, homo2, homo3}. Finally, since DeepPrint is able to encode features that go beyond fingerprint minutiae, it is able to match poor quality fingerprints when reliable minutiae extraction is not possible (Figs.~\ref{fig:intro_fig} and~\ref{fig:feature_maps}).

To arrive at a compact and discriminative representation of only 200 bytes, the DeepPrint architecture is embedded with fingerprint domain knowledge via an automatic alignment module and a multi-task learning objective which requires minutiae-detection (in the form of a \textit{minutiae-map}) as a side task to representation learning. More specifically, DeepPrint automatically aligns an input fingerprint and subsequently extracts both a \textit{texture representation} and a \textit{minutiae-based representation} (both with 96 features). The 192-dimensional concatenation of these two representations, followed by compression from floating point features to integer value features comprises a 200 byte fixed-length representation (192 bytes for the feature vector and 4 bytes for storing the 2 compression parameters). As a final step, we utilize Product Quantization~\cite{product_quant} to further compress the DeepPrint representations stored in the gallery, significantly reducing the computational requirements and time for large-scale fingerprint search.

Detecting minutiae (in the form of a minutiae-map) as a side-task to representation learning has several key benefits:

\begin{itemize}
    \item We guide our representation to incorporate domain inspired features pertaining to minutiae by sharing parameters between the minutiae-map output task and the representation learning task in the multi-task learning framework.
    \item Since minutiae representations are the most popular for fingerprint recognition, we posit that our method for guiding the DeepPrint feature extraction via its minutiae-map side-task falls in line with the goal of ``Explainable AI"~\cite{EAI}.
    \item Given a probe fingerprint, we first use its DeepPrint representation to find the top $k$ candidates and then re-rank the top $k$ candidates using the minutiae-map provided by DeepPrint~\footnote{The $128\times128\times6$ DeepPrint minutiae-map can be easily converted into a minutiae-set with $n$ minutia: $\{(x_1, y_1, \theta_1), ..., (x_n, y_n, \theta_n)\}$ and passed to any minutia-matcher (e.g., COTS A, COTS B, or~\cite{cao}).}. This \textit{optional} re-ranking add-on further improves both accuracy and interpretability.
\end{itemize}

The primary benefit of the 200 byte representation extracted by DeepPrint comes into play when performing mega-scale search against millions or even billions of identities (e.g., India's Aadhaar~\cite{india1} and the FBI's Next Generation Identification (NGI) databases~\cite{ngi}). To highlight the significance of this benefit, we benchmark the search performance of DeepPrint against the latest version SDKs (as of July, 2019) of two top performers in the NIST FpVTE 2012 (Innovatrics\footnote{\url{https://www.innovatrics.com/}} v7.2.1.40 and Verifinger\footnote{\url{https://www.neurotechnology.com/}} v10.0\footnote{We note that Verifinger v10.0 performs significantly better than earlier versions of the SDK often used in the literature.}) on the NIST SD4~\cite{sd4} and NIST SD14~\cite{sd14} databases augmented with a gallery of nearly 1.1 million rolled fingerprints. Our empirical results demonstrate that DeepPrint is competitive with these two state-of-the-art COTS matchers in accuracy while requiring only a fraction of the search time. Furthermore, a given DeepPrint fixed-length representation can also be matched in the encrypted domain via homomorphic encryption with minor loss to recognition accuracy as shown in~\cite{homomorphic} for face recognition. 

\begin{table*}[t]
 \centering
\caption{Published Studies on Fixed-Length Fingerprint Representations}
\label{table:prior_work}
\begin{threeparttable}
\resizebox{\textwidth}{!}{%
\begin{tabular}{ c|c|cccc}
\toprule
Algorithm & Description & \specialcell{HR @ PR = 1.0\%\tnote{1} \\ (NIST SD4)\tnote{2}} & \specialcell{HR @ PR = 1.0\% \\ (NIST SD14)\tnote{3}} & \specialcell{Template Size \\ (bytes)} & \specialcell{Gallery \\Size\tnote{4}} \\
\hline
 Jain~\textit{et al.}~\cite{fingercode, fingercode2} & \specialcell{\textbf{Fingercode}: Global representation \\extracted using Gabor Filters} & N.A. & N.A. & 640 & N.A. \\ 
 \hline
 Cappelli~\textit{et al.}~\cite{mcc} & \specialcell{\textbf{MCC}: Local descriptors via \\ 3D cylindrical structures \\ comprised of the minutiae-set representation} & 93.2\% & 91.0\% & 1,913 & 2,700 \\
 \hline
Cao and Jain~\cite{index1} & \specialcell{\textbf{Inception v3:} Global deep \\ representation extracted via \\Alignment and Inception v3} & 98.65\% & 98.93\% & 8,192 & 250,000 \\
\hline
Song and Feng~\cite{index2} & \specialcell{\textbf{PDC}: Deep representations extracted at \\different resolutions and aggregated\\ into global representation } & 93.3\% & N.A. & N.A. & 2,000 \\
\hline
Song~\textit{et al.}~\cite{index3} & \specialcell{\textbf{MDC}: Deep representations extracted \\from minutiae and aggregated into \\ global representation} & 99.2\% & 99.6\% & 1,200 & 2,700 \\
\hline
Li~\textit{et al.}~\cite{index4} & \specialcell{\textbf{Finger Patches:} Local deep \\representations aggregated into \\global representation via global \\ average pooling} & \textbf{99.83\%} & 99.89\% & 1,024 & 2,700 \\
\hline
Proposed & \specialcell{\textbf{DeepPrint}: Global deep representation \\extracted via multi-task CNN\\ with built-in fingerprint alignment} & 99.75\% & \textbf{99.93\%} & \textbf{200}\tnote{\textdagger} & \textbf{1,100,000} \\
\bottomrule
 
\end{tabular}}
\begin{tablenotes}
\item[1] In some baselines we estimated the data points from a Figure (specific data points were not reported in the paper).
\item[2] Only 2,000 fingerprints are included in the gallery to enable comparison with previous works. (HR = Hit Rate, PR = Penetration Rate)
\item[3] Only last 2,700 pairs (2,700 probes; 2,700 gallery) are used to enable comparison with previous works.
\item[4] Largest gallery size used in the paper.
\item[\textdagger] The DeepPrint representation can be further compressed to only 64 bytes using product quantization with minor loss in accuracy.
\end{tablenotes}
\end{threeparttable}
\end{table*}

More concisely, the primary contributions of this work are:

\begin{itemize}

\item A customized deep network (Fig.~\ref{fig:schematic}), called DeepPrint, which utilizes fingerprint domain knowledge (alignment and minutiae detection) to learn and extract a discriminative fixed-length fingerprint representation.

\item Demonstrating in a manner similar to~\cite{face6} that Product Quantization can be used to compress DeepPrint \textit{fingerprint} representations, enabling even faster mega-scale search (51 ms search time against a gallery of 1.1 million fingerprints vs. 27,000 ms for a COTS with comparable accuracy).

\item Demonstrating with a two-stage search scheme similar to~\cite{face6} that candidates retrieved by DeepPrint representations can be re-ranked using a minutiae-matcher in conjunction with the DeepPrint minutiae-map. This further improves system interpretability and accuracy and demonstrates that the DeepPrint features are complementary to the traditional minutiae representation.

\item Benchmarking DeepPrint against two state-of-the-art COTS matchers (Innovatrics and Verifinger) on NIST SD4 and NIST SD14 against a gallery of 1.1 million fingerprints. Empirical results demonstrate that DeepPrint is comparable to COTS matchers in accuracy at a significantly faster search speed.

\item Benchmarking the authentication performance of DeepPrint on the NIST SD4 and NIST SD14 rolled-fingerprints databases and the FVC 2004 DB1 A slap fingerprint database~\cite{fvc2004}. Again, DeepPrint shows comparable performance against the two COTS matchers, demonstrating the generalization ability of DeepPrint to both rolled and slap fingerprint databases.

\item Demonstrating that homomorphic encryption can be used to match DeepPrint templates in the encrypted domain, in real time (1.26 ms), with minimal loss to matching accuracy as shown for fixed-length face representations~\cite{homomorphic}.

\item An interpretability visualization which demonstrates our ability to guide DeepPrint to look at minutiae-related features.

\end{itemize}

\begin{figure*}[t]
\begin{center}
\includegraphics[scale=0.5]{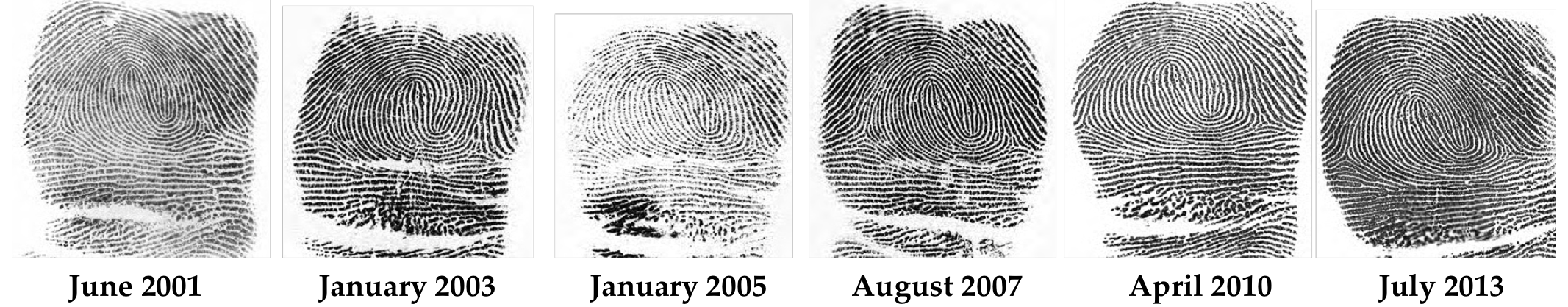}
\caption{Fingerprint impressions from one subject in the DeepPrint training dataset~\cite{longitudinal}. Impressions were captured longitudinally, resulting in the variability across impressions (contrast and intensity from environmental conditions; distortion and alignment from user placement). Importantly, training with longitudinal data enables learning compact representations which are invariant to the typical noise observed across fingerprint impressions over time, a necessity in any fingerprint recognition system.}
\label{fig:longitudinal}
\end{center}
\end{figure*}

\section{Prior Work}

Several early works~\cite{fingercode, fingercode2, mcc} presented fixed-length fingerprint representations using traditional image processing techniques. In~\cite{fingercode, fingercode2}, Jain~\textit{et al.} extracted a global fixed-length representation of 640 bytes, called Fingercode, using a set of Gabor Filters. Cappelli~\textit{et al.} introduced a fixed-length minutiae descriptor, called Minutiae Cylinder Code (MCC), using 3D cylindrical structures computed with minutiae points\cite{mcc}. While both of these representations demonstrated success at the time they were proposed, their accuracy is now significantly inferior to state-of-the-art COTS matchers

Following the seminal contributions of~\cite{fingercode, fingercode2} and~\cite{mcc}, the past 10 years of research on fixed-length fingerprint representations~\cite{f1, f2, f3, f4, f5, f6, f7, f8, f9} has not produced a representation competitive in terms of fingerprint recognition accuracy with the traditional minutiae-based representation. However, recent studies~\cite{index1, index2, index3, index4} have utilized deep networks to extract highly discriminative fixed-length fingerprint representations. More specifically, (i) Cao and Jain~\cite{index1} used global alignment and Inception v3 to learn fixed-length fingerprint representations. (ii) Song and Feng~\cite{index2} used deep networks to extract representations at various resolutions which were then aggregated into a global fixed-length representation. (iii) Song~\textit{et al.}~\cite{index3} further learned fixed-length minutiae descriptors which were aggregated into a global fixed-length representation via an aggregation network. Finally, (v) Li~\textit{et al.}~\cite{index4} extracted local descriptors from predefined ``fingerprint classes" which were then aggregated into a global fixed-length representation through global average pooling.   

While these efforts show tremendous promise, each method has some limitations. In particular, (i) the algorithms proposed in~\cite{index1} and~\cite{index2} both required computationally demanding global alignment as a preprocessing step, and the accuracy is inferior to state-of-the-art COTS matchers. (ii) The representations extracted in~\cite{index3} require the arduous process of minutiae-detection, patch extraction, patch-level inference, and an aggregation network to build a single global feature representation. (iii) While the algorithm in~\cite{index4} obtains high performance on rolled fingerprints (with small gallery size), the accuracy was not reported for slap fingerprints. Since~\cite{index4} aggregates local descriptors by averaging them together, it is unlikely that the approach would work well when areas of the fingerprint are occluded or missing (often times the case in slap fingerprint databases like FVC 2004 DB1 A), and (v) all of the algorithms, suffer from lack of interpretability compared to traditional minutiae representations.  

In addition, existing studies targeting deep, fixed-length fingerprint representations all lack an extensive, large-scale evaluation of the deep features. Indeed, one of the primary motivations for fixed-length fingerprint representations is to perform orders of magnitude faster large scale search. However, with the exception of Cao and Jain~\cite{index1}, who evaluate against a database of 250K fingerprints, the next largest gallery size used in any of the aforementioned studies is only 2,700.

As an addendum, deep networks have also been used to improve \textit{specific sub-modules} of fingerprint recognition systems such as segmentation~\cite{seg1, seg2, seg3, seg4}, orientation field estimation~\cite{orien1,orien2,orien3}, minutiae extraction~\cite{minut1, minut2, minut3}, and minutiae descriptor extraction~\cite{descript1}. However, these works all still operate within the conventional paradigm of extracting an unordered, variable length set of minutiae for fingerprint matching.

\section{DeepPrint}

In the following section, we (i) provide a high-level overview and intuition of DeepPrint, (ii) present how we incorporate automatic alignment into DeepPrint, and (iii) demonstrate how the accuracy and interpretability of DeepPrint is improved through the injection of fingerprint domain knowledge.   

\begin{algorithm}[t]
 
\caption{Extract DeepPrint Representation} \begin{algorithmic}[1]
  \State \textbf{$L(I_f)$}: Shallow localization network, outputs $x,y,\theta$
  \State A: Affine matrix composed with parameters $x,y,\theta$
  \State \textbf{$G(I_f, A)$}: Bilinear grid sampler, outputs aligned fingerprint
  \State \textbf{$S(I_t)$}: Inception v4 stem
  \State \textbf{$E(\mathbf{x})$}: Shared minutiae parameters
  \State \textbf{$M(\mathbf{x})$}: Minutia representation branch
  \State \textbf{$D(\mathbf{x})$}: Minutiae map estimation
  \State \textbf{$T(\mathbf{x})$}: Texture representation branch
\State 
\State \textbf{Input:} Unaligned $448\times448$ fingerprint image $I_f$
\State $A  \gets (x, y, \theta) \gets L(I_f)$ 
\State $I_t \gets G(I_f, A)$
\State $F_{map} \gets S(I_t)$
\State $M_{map} \gets E(F_{map})$
\State $R_{1} \gets M(M_{map})$
\State $H \gets D(M_{map})$
\State $R_{2} \gets T(F_{map})$
\State $\mathbf{R} \gets R_{1} \oplus R_{2}$
\State \textbf{Output:} Fingerprint representation $\mathbf{R} \in \mathbb{R}^{192}$ and minutiae-map $H$. ($H$ can be optionally utilized for (i) visualization and (ii) fusion of DeepPrint scores obtained via $\mathbf{R}$ with minutiae-matching scores.)
\end{algorithmic} 
\label{alg:alg1}
\end{algorithm} 

\begin{figure*}
\begin{minipage}{.2\textwidth}
\centering
\includegraphics[scale=.2]{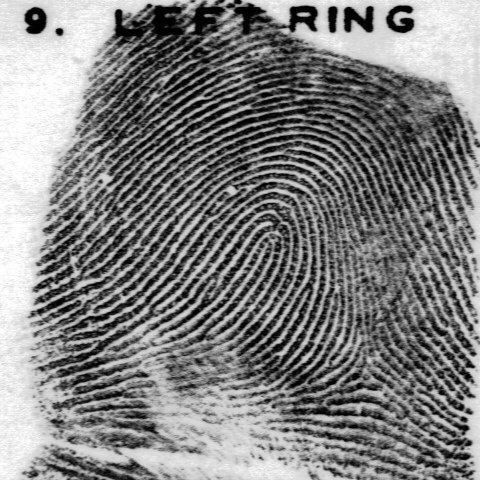}
\end{minipage}%
\begin{minipage}{.2\textwidth}
\centering
\includegraphics[scale=.2]{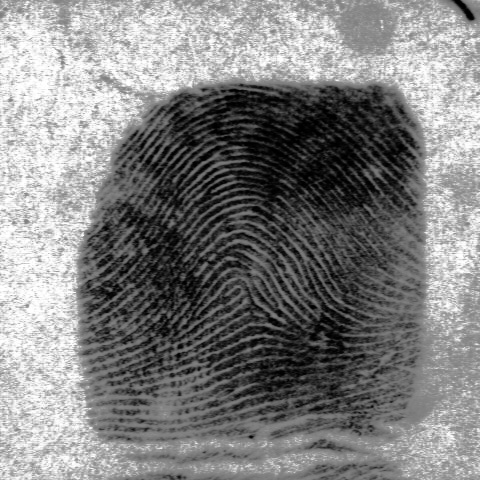}
\end{minipage}
\begin{minipage}{.2\textwidth}
\centering
\includegraphics[scale=.2]{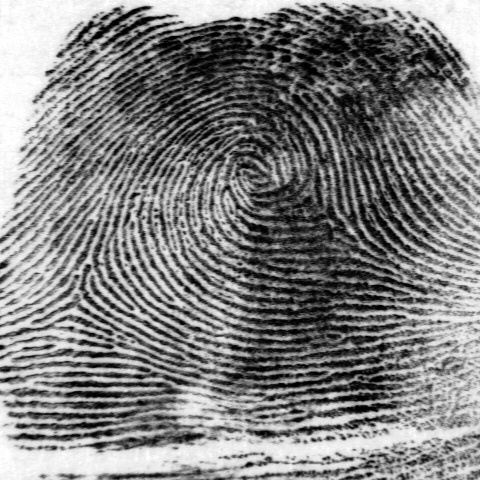}
\end{minipage}%
\begin{minipage}{.2\textwidth}
\centering
\includegraphics[scale=.2]{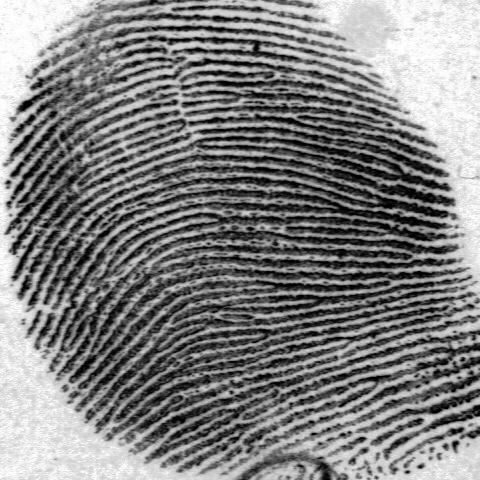}
\end{minipage}%
\begin{minipage}{.2\textwidth}
\centering
\includegraphics[scale=.2]{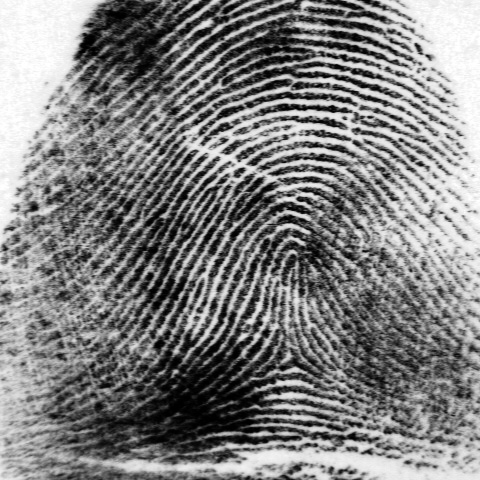}
\end{minipage}

\par\medskip

\begin{minipage}{.2\textwidth}
\centering
\includegraphics[scale=.215]{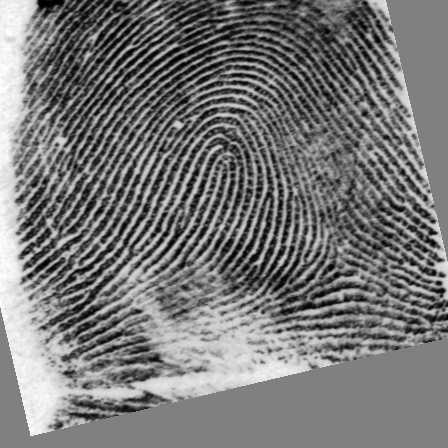}
\end{minipage}%
\begin{minipage}{.2\textwidth}
\centering
\includegraphics[scale=.215]{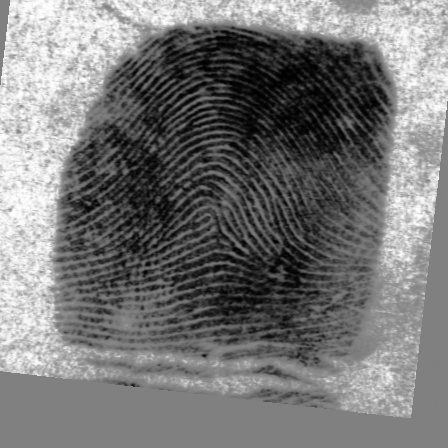}
\end{minipage}
\begin{minipage}{.2\textwidth}
\centering
\includegraphics[scale=.215]{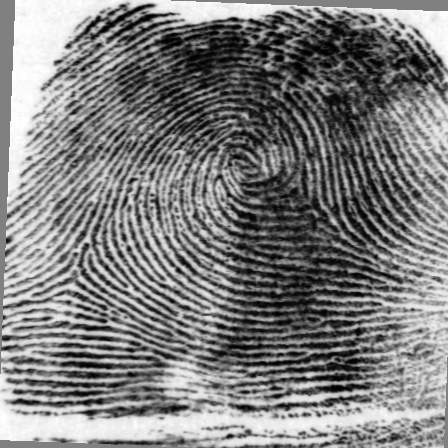}
\end{minipage}%
\begin{minipage}{.2\textwidth}
\centering
\includegraphics[scale=.215]{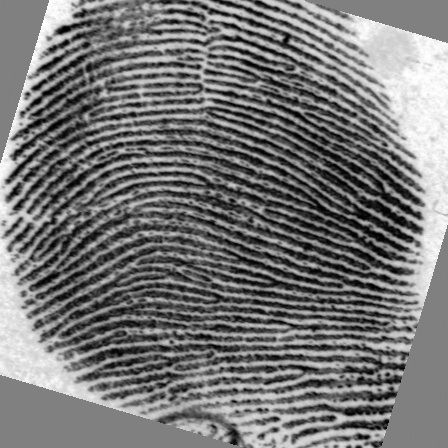}
\end{minipage}%
\begin{minipage}{.2\textwidth}
\centering
\includegraphics[scale=.215]{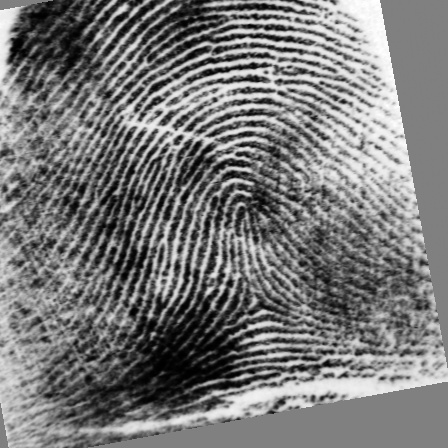}
\end{minipage}%
\centering
\caption{Unaligned fingerprint images from NIST SD4 (top row) and corresponding DeepPrint aligned fingerprint images (bottom row).}
\label{fig:transform}
\end{figure*}

\subsection{Overview}

A high level overview of DeepPrint is provided in Figure~\ref{fig:schematic} with pseudocode in Algorithm~\ref{alg:alg1}. DeepPrint is trained with a longitudinal database (Fig.~\ref{fig:longitudinal}) comprised of 455K rolled fingerprint images stemming from 38,291 unique fingers~\cite{longitudinal}. Longitudinal fingerprint databases consist of fingerprints from distinct subjects captured over time (Fig.~\ref{fig:longitudinal})~\cite{longitudinal}. It is necessary to train DeepPrint with a large, longitudinal database so that it can learn compact, fixed-length representations which are invariant to the differences introduced during fingerprint image acquisition at different times and in different environments (humidity, temperature, user interaction with the reader, and finger injuries). The primary task during training is to predict the finger identity label $c \in [0,38291]$ (encoded as a one-hot vector) of each of the 455K training fingerprint images ($\approx12$ fingerprint impressions / finger). The last fully connected layer is taken as the representation for fingerprint comparison during authentication and search.

The input to DeepPrint is a $448\times448$~\footnote{Fingerprint images in our training dataset vary in size from $\approx~512~\times~512$ to $\approx~800~\times~800$. As a pre-processing step, we do a center cropping (using Gaussian filtering, dilation and erosion, and thresholding) to all images to $\approx~448~\times~448$. This size is sufficient to cover most of the rolled fingerprint area without extraneous background pixels.} grayscale fingerprint image, $I_f$, which is first passed through the alignment module (Fig.~\ref{fig:schematic}). The alignment module consists of a localization network, $L$, and a grid sampler, $G$~\cite{spatial}. After applying the localization network and grid sampler to $I_f$, an aligned fingerprint $I_t$ is passed to the base-network, $S$. 

The base-network is the stem of the Inception v4 architecture (Inception v4 minus Inception modules). Following the base-network are two different branches (Fig.~\ref{fig:schematic}) comprised primarily of the three Inception modules (A, B, and C) described in~\cite{inceptionv4}. The first branch, $T(x)$, completes the Inception v4 architecture~\footnote{We selected Inception v4 after evaluating numerous other architectures such as: ResNet, Inception v3, Inception ResNet, and MobileNet.} as $T(S(I_t))$ and performs the primary learning task of predicting a finger identity label directly from the cropped, aligned fingerprint $I_t$. It is included in order to learn the textural cues in the fingerprint image. The second branch (Figs.~\ref{fig:schematic} and~\ref{fig:minutiae_arch}), $M(E(S(I_t)))$, again predicts the finger identity label from the aligned fingerprint $I_t$, but it also has a related side task (Fig.~\ref{fig:minutiae_arch}) of detecting the minutiae locations and orientations in $I_t$ via $D(E(S(I_t)))$. In this manner, we guide this branch of the network to extract representations influenced by fingerprint minutiae (since parameters between the minutiae detection task and representation learning task are shared in $E(x)$). The textural cues act as complementary discriminative information to the minutiae-guided representation. The two 96-dimensional representations (each dimension is a float, consuming 4 bytes of space) are concatenated into a 192-dimensional representation (768 total bytes). Finally, the floats are truncated from 32 bits to 8 bit integer values, compressing the template size to 200 bytes (192 bytes for features and 8 bytes for 2 decompression parameters). Note that the minutiae set is not explicitly used in the final representation. Rather, we use the minutiae-map to guide our network training. However, for improved accuracy and interpretability, we can \textit{optionally} store the minutiae set for use in a re-ranking scheme during large-scale search operations. 

In the following subsections, we provide details of the major sub-components of the proposed network architecture.

\subsection{Alignment}

In nearly all fingerprint recognition systems, the first step is to perform alignment based on some reference points (such as the core point). However, this alignment is computationally expensive. This motivated us to adopt attention mechanisms such as the spatial transformers in~\cite{spatial}. 

The advantages of using the spatial transformer module in place of reference point based alignment algorithms are two-fold: (i) it requires only one forward pass through a shallow localization network (Table~\ref{tabel:localization}), followed by bilinear grid sampling. This reduces the computational complexity of alignment (we resize the $448~\times~448$ fingerprints to $128~\times~128$\footnote{We also tried $64~\times~64$, however, we could not obtain consistent alignment at this resolution.} to further speed up the localization estimation); (ii) The parameters of the localization network are tuned to minimize the loss (Eq.~\ref{eq:loss}) of the base-network and representation extraction networks. In other words, rather than supervising the transformation via reference points (such as the core point), we let the base-network and representation extraction networks tell the localization network what a ``good" transformation is, so that it can learn a more discriminative representation for the input fingerprint.

\begin{figure}[t]
\begin{center}
\includegraphics[scale=0.6]{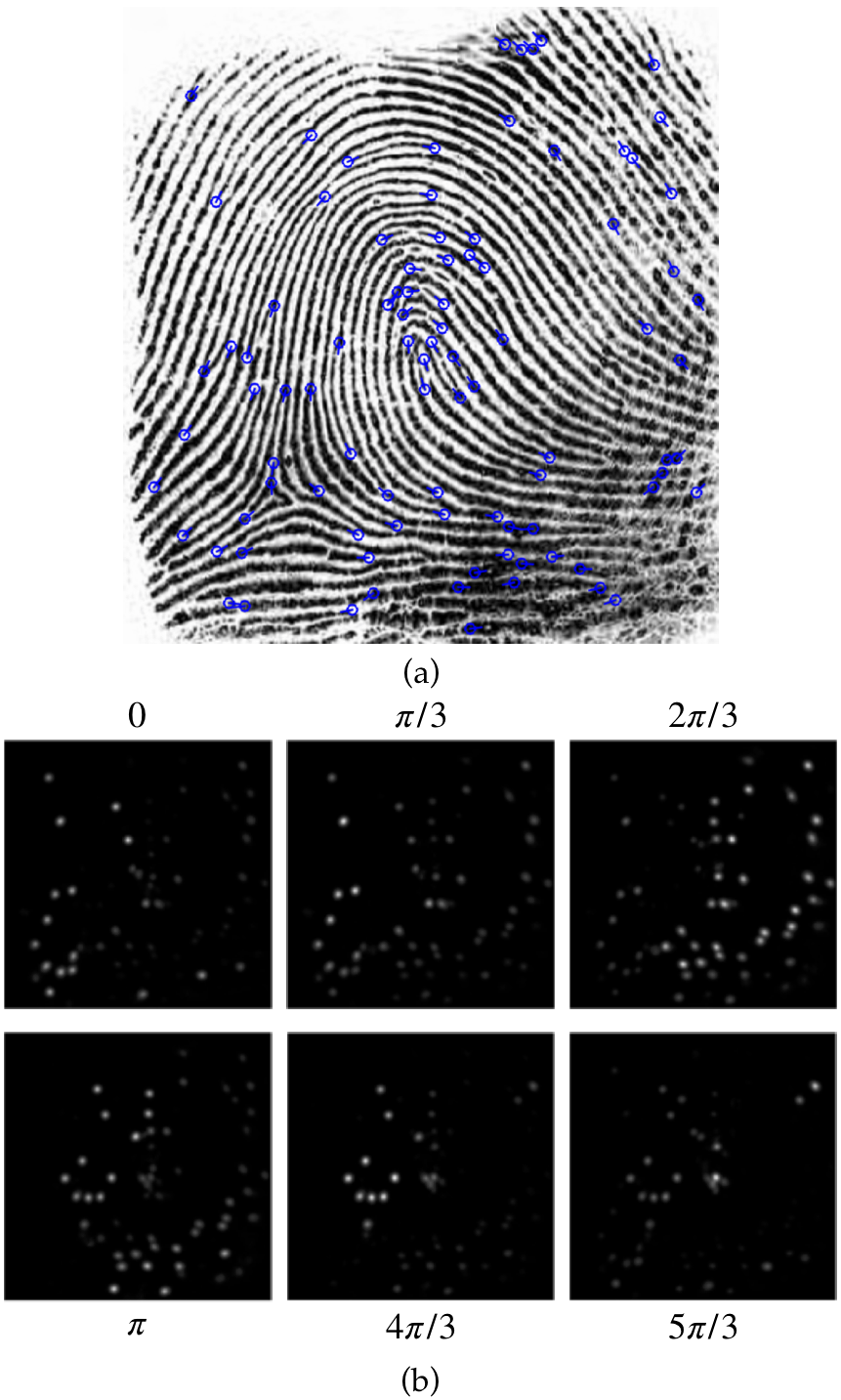}
\caption{Minutiae Map Extraction. The minutiae locations and orientations of an input fingerprint (a) are encoded as a 6-channel minutiae map (b). The ``hot spots" in each channel indicate the spatial location of the minutiae points. The $k^{th}$ channel of the hot spots indicate the contributions of each minutiae to the $k\pi/3$ orientation.}
\label{fig:mmap}
\end{center}
\vspace{-1.5em}
\end{figure} 

\begin{table}[t]
\caption{Localization Network Architecture}
 \centering
\begin{threeparttable}
\begin{tabular}{c c c}
 \toprule
 Type & \specialcell{Output \\ Size} & \specialcell{Filter \\ Size, Stride}\\
 \midrule
 Convolution & $128\times128\times24$ & $5\times5$, $1$\\
 \midrule
 Max Pooling & $64\times64\times24$ & $2\times2$, $2$\\
 \midrule
 Convolution & $64\times64\times32$ & $3\times3$, $1$\\
 \midrule
 Max Pooling & $32\times32\times32$ & $2\times2$, $2$\\
 \midrule
 Convolution& $32\times32\times48$ & $3\times3$, $1$\\
 \midrule
 Max Pooling & $16\times16\times48$ & $2\times2$, $2$\\
 \midrule
 Convolution& $16\times16\times64$ & $3\times3$, $1$\\
 \midrule
 Max Pooling & $8\times8\times64$ & $2\times2$, $2$\\
\midrule
 Fully Connected & $64$ &  \\
 \midrule
  Fully Connected & $3$\tnote{\textdagger} &  \\
 \bottomrule
\end{tabular}
\begin{tablenotes}
\item[\textdagger] These three outputs correspond to $x$,$y$,$\theta$ shown in Fig.~\ref{fig:schematic}.
\end{tablenotes}
\end{threeparttable}
\label{tabel:localization}
\end{table}

Given an unaligned fingerprint image $I_f$, a shallow localization network first hypothesizes the translation and rotation parameters ($x$,$y$, and $\theta$) of an affine transformation matrix $A_\theta$ (Fig.~\ref{fig:schematic}). A user specified scaling parameter $\lambda$ is used to complete $A_\theta$ (Fig.~\ref{fig:schematic}). This scaling parameter stipulates the area of the input fingerprint image which will be cropped. We train two DeepPrint models, one for rolled fingerprints ($\lambda=1$) and one for slap fingerprints ($\lambda=\frac{285}{448}$) meaning a $285~\times~285$ fingerprint area window will be cropped from the $448~\times~448$ input fingerprint image. Given $A_\theta$, a grid sampler $G$ samples the input image $I_f$ pixels $(x_i^f, y_i^f)$ for every target grid location $(x_i^t, y_i^t)$ to output the aligned fingerprint image $I_t$ in accordance with Equation~\ref{eq:alignment}.

\begin{equation}
\begin{pmatrix}x_i^f\\ y_i^f\\ 1\end{pmatrix} = A_\theta\begin{pmatrix}x_i^t\\ y_i^t\\ 1\end{pmatrix}
\label{eq:alignment}
\end{equation} Once $I_t$ has been computed, it is passed on to the base-network for classification. Finally, the parameters for the localization network are updated based upon the loss in Equation~\ref{eq:loss}.

The architecture used for our localization network is shown in Table~\ref{tabel:localization} and images from before and after the alignment module are shown in Figure~\ref{fig:transform}. In order to get the localization network to properly converge, (i) the learning rate was scaled by $0.035$ and (ii) the upper bound of the estimated affine matrix translation and rotation parameters was set to $224$ pixels and $\pm 60$ degrees, respectively. These constraints are based on our domain knowledge on the maximum extent a user would rotate or translate their fingers during placement on the reader platen.

\subsection{Minutiae Map Domain Knowledge}

To prevent overfitting the network to the training data and to extract interpretable deep features, we incorporate fingerprint domain knowledge into DeepPrint. The specific domain knowledge we incorporate into our network architecture is hereafter referred to as the \textit{minutiae map}~\cite{cao}. Note that the minutiae map is not explicitly used in the fixed-length fingerprint representation, but the information contained in the map is indirectly embedded in the network during training.

A minutiae map is essentially a 6-channel heatmap quantizing the locations $(x, y)$ and orientations $\theta \in [0,2\pi]$ of the minutiae within a fingerprint image. More formally, let $h$ and $w$ be the height and width of an input fingerprint image and $T = \{m_1, m_2, ..., m_n\}$ be its minutiae template with $n$ minutiae points, where $m_t = (x_t, y_t, \theta_t)$ and $t = 1,...,n$. Then, the minutiae map $H \in \mathbb{R}^{h\times w\times 6}$ at $(i, j, k)$ can be computed by summing the location and orientation contributions of each of the minutiae in $T$ to obtain the heat map (Fig.~\ref{fig:mmap} (b)).

\begin{equation}
H(i, j, k) = \sum_{t=1}^n C_s((x_t,y_t), (i, j)) \cdot C_o(\theta_t, 2k\pi /6)
\end{equation} where $C_s(.)$ and $C_o(.)$ calculate the spatial and orientation contribution of minutiae $m_t$ to the minutiae map at $(i, j, k)$ based upon the euclidean distance of $(x_t,y_t)$ to $(i,j)$ and the orientation difference between $\theta_t$ and $2k\pi /6$ as follows:

\begin{equation}
C_s((x_t,y_t), (i,j)) = exp(-\frac{||(x_t,y_t)-(i,j)||_2^2}{2\sigma_s^2})
\end{equation}

\begin{equation}
C_o(\theta_t,2k\pi /6) = exp(-\frac{d\phi(\theta_t,2k\pi /6)}{2\sigma_s^2})
\end{equation} where $\sigma_s^2$ is the parameter which controls the width of the gaussian, and $d\phi(\theta_1,\theta_2)$ is the orientation difference between angles $\theta_1$ and $\theta_2$:

\begin{equation}
d\phi(\theta_1, \theta_2) = \begin{cases}
|\theta_1 - \theta_2| & -\pi \leq \theta_1 - \theta_2 \leq \pi \\
2\pi - |\theta_1 - \theta_2| & otherwise.
\end{cases}
\end{equation} An example fingerprint image and its corresponding minutiae map are shown in Figure~\ref{fig:mmap}. A minutiae-map can be converted back to a minutiae set by finding the local maximums in a channel (location), and individual channel contributions (orientation), followed by non-maximal suppression to remove spurious minutiae\footnote{ Code for converting a minutiae set to a minutiae map and vice versa is open-sourced:~\url{https://bit.ly/2KpbPxV}}.

\subsection{Multi-Task Architecture}

\begin{figure}[t]
\begin{center}
\includegraphics[scale=0.5]{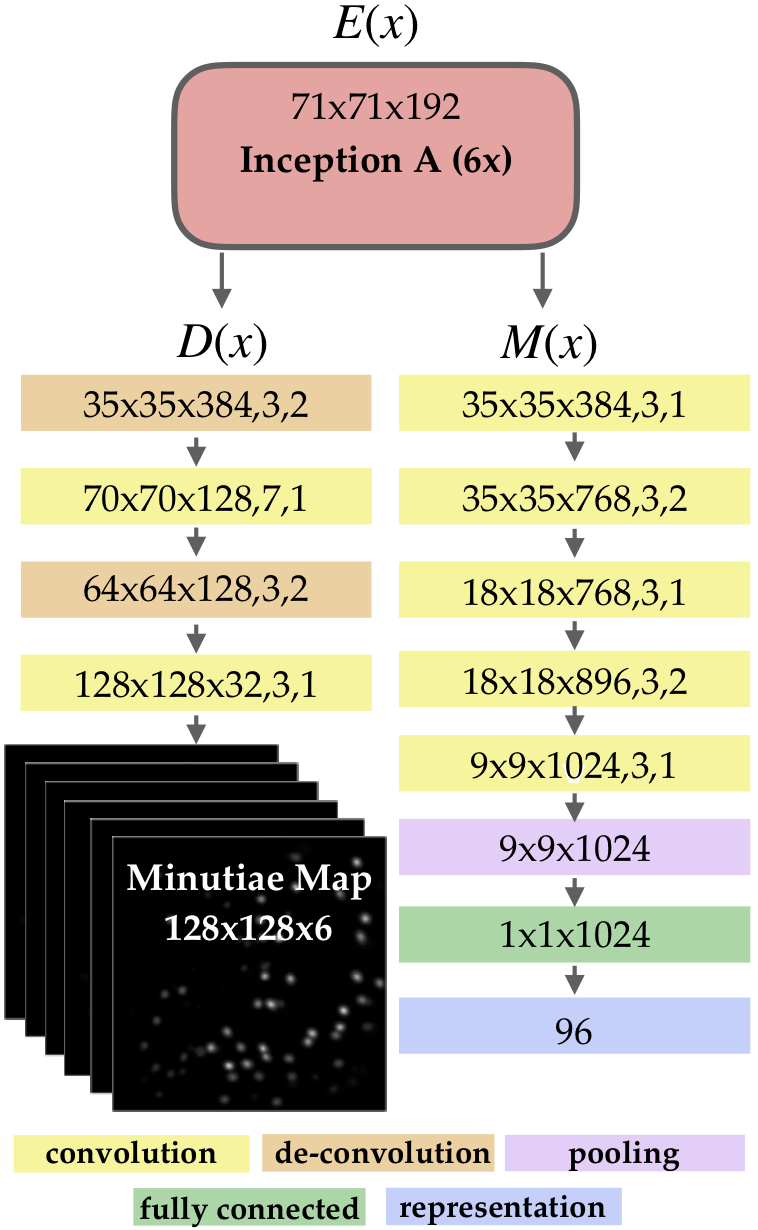}
\caption{The custom multi-task \textbf{minutiae branch} of DeepPrint. The dimensions inside each box represent the input dimensions, kernel size, and stride length, respectively.}
\label{fig:minutiae_arch}
\end{center}
\vspace{-1.0em}
\end{figure}

The minutiae-map domain knowledge is injected into DeepPrint via multitask learning. Multitask learning improves generalizability of a model since domain knowledge within the training signals of related tasks acts as an inductive bias~\cite{multi, multi0}. The multi-task branch of the DeepPrint architecture is shown in Figures~\ref{fig:schematic} and~\ref{fig:minutiae_arch}. The primary task of the branch is to extract a representation and subsequently classify a given fingerprint image into its ``finger identity". The secondary task is to estimate the minutiae-map. Since parameters are shared between the representation learning task and the minutiae-map extraction task, we guide the minutiae-branch of our network to extract fingerprint representations that are influenced by minutiae locations and orientations. At the same time, a separate branch in DeepPrint aims to extract a complementary texture-based representation by directly predicting the identity of an input fingerprint without any domain knowledge (Fig.~\ref{fig:schematic}). DeepPrint extracts minutiae maps of size $128\times128\times6$~\footnote{We extract maps of $128\times128\times6$ to save GPU memory during training (enabling a larger batch size), and to reduce disk space requirements for storage of the maps.} to encode the minutiae locations and orientations of an input fingerprint image of size $448\times448\times1$. The ground truth minutiae maps for training DeepPrint are estimated using the open source minutiae extractor proposed in~\cite{cao}.

Note, we combine the texture branch with the minutiae branch in the DeepPrint architecture (rather than two separate networks) for the following reasons: (i) the minutiae branch and the texture branch share a number of parameters (the Inception v4 stem), reducing the model complexity that two separate models would necessitate, and (ii) the spatial transformer (alignment module) is optimized based on both branches (\textit{i.e.} learned alignment benefits both the texture-based and minutiae-based representations) avoiding two separate spatial transformer modules and alignments.

More formally, we incorporate domain knowledge into the DeepPrint representation by computing the network's loss in the following manner. First, given $R_1$ and $R_2$ as computed in Algorithm~\ref{alg:alg1}, fully connected layers are applied for identity classification logits, outputting $\mathbf{y}_1 \in \mathbb{R}^{c}$ and $\mathbf{y}_2 \in \mathbb{R}^{c}$, where $c$ is the number of identities in the training set. Next, $\mathbf{y}_1$ and $\mathbf{y}_2$ are both passed to a softmax layer to compute the probabilities $\mathbf{\hat{y}}_1$ and $\mathbf{\hat{y}}_2$ of $R_1$ and $R_2$ belonging to each identity. Finally, $\mathbf{\hat{y}}_1$ and $\mathbf{\hat{y}}_2$, the ground truth label $y$, and the network's parameters $w$, can be used to compute the combined cross-entropy loss of the two branches and an image $I_t$:

\begin{equation}
\mathcal{L}_1(I_t, y) = -log(\mathbf{\hat{y}}_1^{j=y} | I_t, w) -log(\mathbf{\hat{y}}_2^{j=y} | I_t, w)
\end{equation}
where $~j \in \{1, \cdots, c\}$. To further reduce the intra-class variance of the learned features, we also employ the widely used center-loss first proposed in~\cite{centerloss} for face recognition. In particular, we compute two center-loss terms, one for each branch in our multi-task architecture as:

\begin{equation}
\mathcal{L}_2(I_t) =  ||R_1 - ctr_{1}^n||_2^2 + ||R_2 - ctr_{2}^n||_2^2
\end{equation} where $ctr_{i}^n$, are the branch, $i$, and subject, $n$, specific centers for a fingerprint image $I_t$.

For computing the loss of the minutiae map estimation side task, we employ the Mean Squared Error Loss between the estimated minutiae map $\textbf{H}$ and the ground truth minutiae map~\footnote{The ground truth minutiae maps are estimated using the open-source minutiae extractor in~\cite{cao}.} $H$ as follows:

\begin{equation}
\mathcal{L}_3(I_t, H) = \sum_{i,j,k} (\textbf{H}_{i,j,k} - H_{i,j,k})^2
\end{equation} 

Finally, using the addition of all these loss terms, and a dataset comprised of $N$ training images, our model parameters $w$ are trained in accordance with:

\begin{multline}
\argmin_w \sum_{i=1}^N \lambda_1\mathcal{L}_1(I_t^i, y^i)~+~\lambda_2\mathcal{L}_2(I_t^i)~+~\lambda_3\mathcal{L}_3(I_t^i,H^i)
\label{eq:loss}
\end{multline} where $\{\lambda_1=1, \lambda_2=0.00125, \lambda_3=0.095\}$ are empirically set to obtain convergence. Note, during the training, we augment our dataset with random rotations, translations, brightness, and cropping. We use the RMSProp optimizer with a batch size of 30. Weights are initialized with the variance scaling initializer. Regularization included dropout (before the embedding fully connected layer) with a keep probability of $0.8$ and weight decay of $0.00004$. We trained for 140K steps, which lasted 25 hours. 

After the multitask architecture has converged, a fixed length feature representation can be acquired by extracting the fully connected layer before the softmax layers in both of the network's branches. Let $R_1 \in \mathbb{R}^{96}$ be the unit-length minutiae representation and $R_2 \in \mathbb{R}^{96}$ be the unit-length texture representation. Then, a final feature representation is obtained by concatenation of $R_1$ and $R_2$ into $\mathbf{R} \in \mathbb{R}^{192}$, followed by normalization of $\mathbf{R}$ to unit length. 

\subsection{Template Compression}

The final step in the DeepPrint representation extraction is template compression. In particular, the 192-dimensional DeepPrint representation consumes a total of 768 bytes. We can compress this size to 200 bytes by truncating the 32 bit floating point feature values to 8-bit integer values in the range of [0,255] using min-max normalization. In particular, given a DeepPrint representation $\mathbf{R}\in~\mathbb{R}^{192}$, we transfer the domain of $\mathbf{R}$ to $\mathbf{R}\in~\mathbb{N}^{192}$ and output $\mathbf{R}^{\prime}$, where we restrict the set of the natural numbers $\mathbb{N}$ to the range of [0,255]. More formally:

\begin{equation}
    \mathbf{R}^{\prime} = \floor*{\frac{255(\mathbf{R} - min(\mathbf{R}))}{max(\mathbf{R}) - min(\mathbf{R})}}
\end{equation} where $min(\mathbf{R})$ and $max(\mathbf{R})$ output the minimum and maximum feature values of the vector $\mathbf{R}$, respectively. In order to decompress the features back to float values for matching, we need to save the minimum and maximum values for each representation. Thus, our final representation is 200 bytes, 192 bytes for the features, 4 bytes for the minimum value and 4 bytes for the maximum value. To decompress the representations (when loading them into RAM), we simply reverse the min-max normalization using the saved minimum and maximum values. Table~\ref{table:compression_effects} shows that compression only minimally impacts the matching accuracy.  

\begin{table}[h]
\begin{threeparttable}   
\caption{Effect of Compression on Accuracy}
\label{table:compression_effects}
\begin{center}
\begin{tabular}{ ccc }
\toprule
Dataset & \specialcell{DeepPrint \\Uncompressed Features} & \specialcell{DeepPrint \\Compressed Features} \\
\midrule
NIST SD4\tnote{\textdagger} & 97.95\% & 97.90\% \\
\midrule
FVC 2004 DB1 A\tnote{\textdagger\textdagger} & 97.53\% & 97.50\% \\
\bottomrule
\end{tabular}
\begin{tablenotes}
\item [\textdagger] TAR @ FAR = 0.01\% is reported.
\item [\textdagger\textdagger] TAR @ FAR = 0.1\% is reported.
\end{tablenotes}
\end{center}
\end{threeparttable}
\end{table}

\section{DeepPrint Matching}

Two, unit length, DeepPrint representations $\mathbf{R}_p$ and $\mathbf{R}_g$ can be easily matched using the cosine similarity between the two representations. In particular:

\begin{equation}
s(\mathbf{R}_p, \mathbf{R}_g) = \mathbf{R}_p^{\intercal} \cdot \mathbf{R}_g
\label{eq:cosine}
\end{equation}

Thus, DeepPrint authentication (1:1 matching) requires only 192 multiplications and 191 additions. We also experimented with euclidian distance as a scoring function, but consistently obtained higher performance with cosine similarity. Note that if compression is added, there would be an additional $d$ subtractions and $d$ multiplications to reverse the min-max normalization of the enrolled representation. Therefore, the authentication time effectively doubles. However, depending on the application or implementation, compression does not necessarily effect the search speed since the gallery of representations could be already decompressed and in RAM before performing a search.

\subsection{Fusion of DeepPrint Score with Minutiae Score}

Given the speed of matching two DeepPrint representations, the minutiae-based match scores of any existing AFIS can also be fused together with the DeepPrint scores with minimal loss to the overall AFIS authentication speed (\textit{i.e.} DeepPrint can be easily used as an add-on to existing minutiae-based AFIS to improve recognition accuracy). In our experimental analysis, we demonstrate this by fusing DeepPrint scores together with the scores of minutiae-based matchers COTS A, COTS B, and~\cite{cao} and subsequently improving authentication accuracy. This indicates that the information contained in the compact DeepPrint representation is complementary to that of minutiae representations. Note, since DeepPrint already extracts minutiae as a side task, fusion with a minutiae-based matcher requires little extra computational overhead (simply feed the minutiae extracted by DeepPrint directly to the minutiae matcher, eliminating the need to extract minutiae a second time).

\section{DeepPrint Search}

Fingerprint search entails finding the top $k$ candidates, in a database (gallery or background) of $N$ fingerprints, for an input probe fingerprint. The simplest algorithm for obtaining the top $k$ candidates is to (i) compute a similarity measure between the probe template and every enrolled template in the database, (ii) sort the enrolled templates by their similarity to the probe~\footnote{In our search experiments, we reduce the typical sorting time from $Nlog(N)$ to $Nlog(k)$ (where $k<<N$) by maintaining a priority queue of size k since we only care about the scores of the top $k$ candidates. This trick reduces sorting time from 23 seconds to 8 seconds when the gallery size $N = 100,000,000$ and the candidate list size $k = 100$.}, and (iii) select the top $k$ most similar enrollees. More formally, finding the top $k$ candidates $C_k(.)$ in a gallery $G$ for a probe fingerprint $\mathbf{R}_p$ is formulated as:

\begin{equation}
    C_k(\mathbf{R}_p) = Rank_k(\{s(\mathbf{R}_p, \mathbf{R}_g) | \mathbf{R}_g \in G\})
    \label{eq:rank}
\end{equation} where $Rank_k(.)$ returns the $k$ most similar candidates from an input set of candidates and $s$ is a similarity function such as defined in Equation~\ref{eq:cosine}.

Since minutiae-based matching is computationally expensive, comparing the probe to every template enrolled in the database in a timely manner is not feasible with minutiae matchers. This has led to a number of schemes to either significantly reduce the search space, or utilize high-level features to quickly index top candidates~\cite{old_index1, old_index2, old_index3, old_index4, old_index5}. However, such methods have not achieved high-levels of accuracy on public benchmark datasets such as NIST SD4 or NIST SD14.

In contrast to minutiae-matchers, the fixed-length, 200 byte DeepPrint representations can be matched extremely quickly using Equation~\ref{eq:cosine}. Therefore, large scale search with DeepPrint can be performed by \textit{exhaustive} comparison of the probe template to every gallery template in accordance with Equation~\ref{eq:rank}. The complexity of exhaustive search is linear with respect to both the gallery size $N$ and the dimensionality $d$ of the DeepPrint representation ($d = 192$ in this case).

\subsection{Faster Search}

Although exhaustive search can be effectively utilized with DeepPrint representations in conjunction with Equation~\ref{eq:rank}, it may be desirable to even further decrease the search time. For example, when searching against 100 million fingerprints, the DeepPrint search time is still (11 seconds on an i9 processor with 64 GB of RAM)~\footnote{Search time for 100 million gallery was simulated by generating 100 million random representations, where each feature was a 32-bit float value drawn from a uniform distribution from 0 to 1.}. A natural way to reduce the search time further with minimal loss to accuracy is to utilize an effective approximate nearest neighbor (ANN) algorithm.

\textit{Product Quantization} is one such ANN algorithm which has been successfully utilized in large-scale face search~\cite{face6}. Product quantization is still an exhaustive search algorithm, however, representations are first compressed via keys to a lookup table, which significantly reduces the comparison time between two representations. In other words, product quantization reformulates the comparison function in Equation~\ref{eq:cosine} to a series of lookup operations in a table stored in RAM. More formally, a given DeepPrint representation $\mathbf{R}_g$ of dimensionality $d$, is first decomposed into $m$ sub-vectors as:

\begin{equation}
    \mathbf{R}_g = (R^1, R^2, ..., R^m)
\end{equation}

Next, each $m^{th}$ sub-vector $R^i \in \mathbb{R}^{d/m}$ is mapped to a codeword $c_j^i$ in a codebook $\mathcal{C}^i = \{c_{j=1, 2, ...,z}^i | c_j^i \in \mathbb{R}^{d / m}\}$ where $z$ is the size of the codebook. The index $j$ of each codeword $c_j^i$ can be represented as a binary code of $log_2(z)$ bits. Therefore, after mapping each sub-vector to its codeword, the original d-dimensional representation $\mathbf{R}_g$ ($d=192$ for DeepPrint) can be compressed to only $m*log_2(z)$ bits!

The codewords $c_j^i \in \mathbb{R}^{d/m}$ for each codebook $\mathcal{C}^i$ are computed offline (before search time) using k-means clustering for each sub-vector. Thus each codebook $\mathcal{C}^i$ contains $z$ centroids computed from the corresponding sub-vectors $R^i$. Given all $m$ codebooks $\{\mathcal{C}^1, \mathcal{C}^2, \mathcal{C}^3, ..., \mathcal{C}^m\}$, the product quantizer of $\mathbf{R}_g$ is computed as:

\begin{equation}
    q(\mathbf{R}_g) = (q^1(R^1), ..., q^m(R^m))
\end{equation} where $q^i(R^i)$ is the index of the nearest centroid in the codebook $\mathcal{C}^i, i=1,...,m$.

Finally, given a DeepPrint probe representation $\mathbf{R}_p$, and the now quantized gallery template $\mathbf{R}_g$, a match score can be obtained in accordance with Equation~\ref{eq:product_match}:

\begin{equation}
    s(\mathbf{R}_p, \mathbf{R}_g) = ||\mathbf{R}_p - q(\mathbf{R}_g)||_2^2 = \sum_{i=1}^m ||\mathbf{R}_p^i - q^i(R^i)||_2^2
    \label{eq:product_match}
\end{equation}

Thus matching a probe template to each quantized template in the gallery requires a one-time build up of a $m~\times~z$ table which is stored in RAM, followed by $m$ lookups and additions for each quantized template in the gallery. In our experiments, we set $z = 256$ and $m = 64$. A quantized template in the gallery is compressed to 64 bytes, and search is reduced from 192 additions and multiplications ($N$ times, where $N$ is the gallery size) to a one-time $m~\times~z$ table build up, followed by 64 lookups and additions for each gallery template (a significant savings on memory and search time)\footnote{We used the Facebook Faiss PQ implementation: \url{https://github.com/facebookresearch/faiss}}.

\begin{figure*}[t]
\begin{minipage}{.2\textwidth}
\centering
\includegraphics[scale=.2]{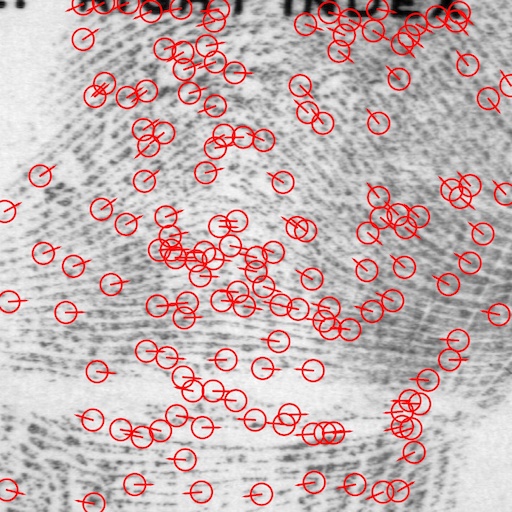}
\end{minipage}%
\begin{minipage}{.2\textwidth}
\centering
\includegraphics[scale=.2]{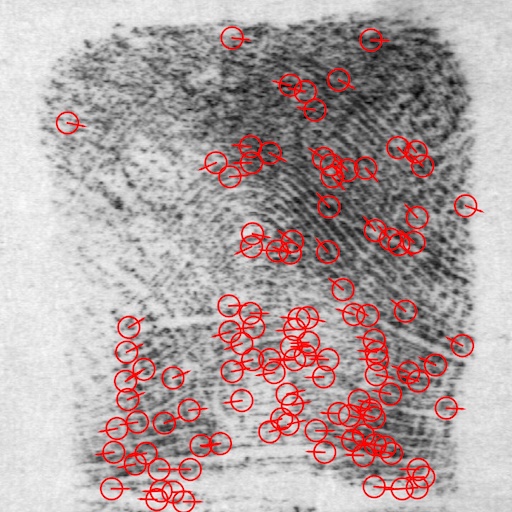}
\end{minipage}
\begin{minipage}{.2\textwidth}
\centering
\includegraphics[scale=.2]{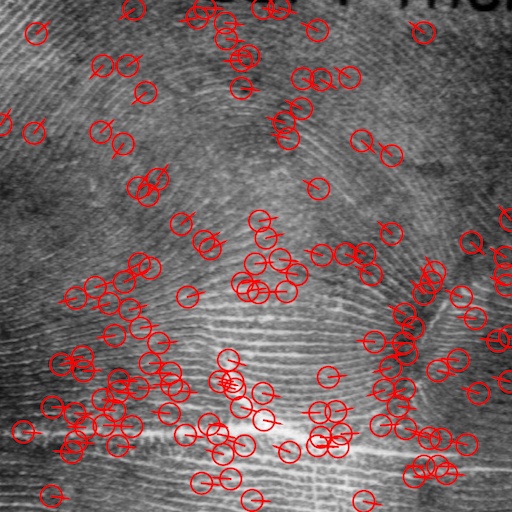}
\end{minipage}%
\begin{minipage}{.2\textwidth}
\centering
\includegraphics[scale=.2]{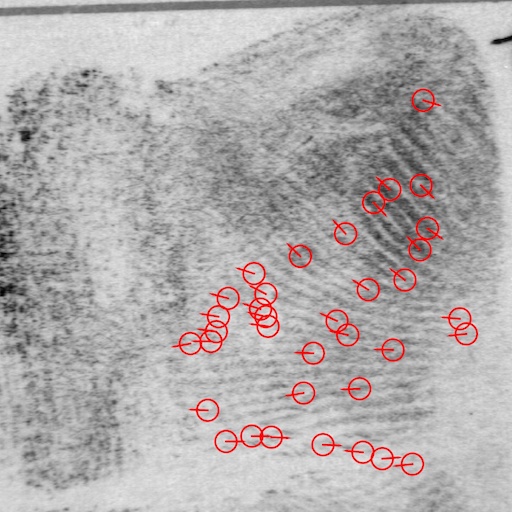}
\end{minipage}%
\begin{minipage}{.2\textwidth}
\centering
\includegraphics[scale=.2]{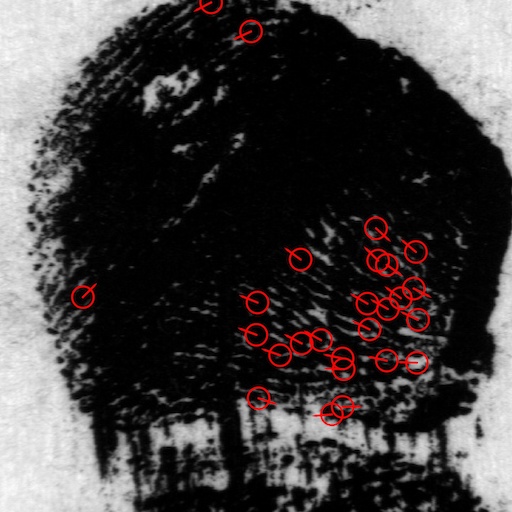}
\end{minipage}

\par\medskip

\begin{minipage}{.2\textwidth}
\centering
\includegraphics[scale=.215]{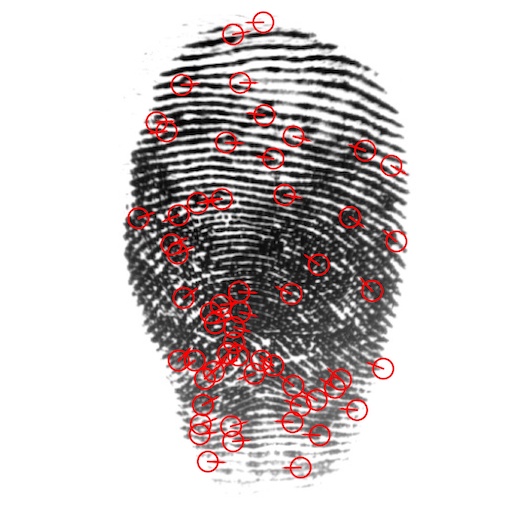}
\end{minipage}%
\begin{minipage}{.2\textwidth}
\centering
\includegraphics[scale=.215]{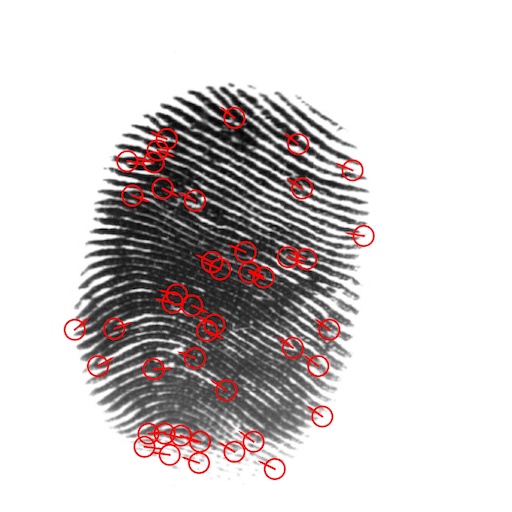}
\end{minipage}
\begin{minipage}{.2\textwidth}
\centering
\includegraphics[scale=.215]{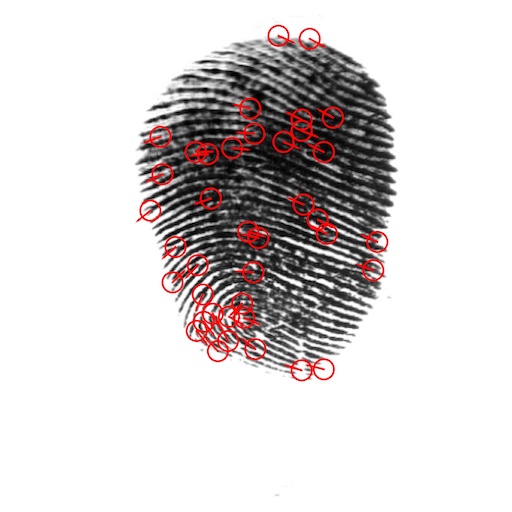}
\end{minipage}%
\begin{minipage}{.2\textwidth}
\centering
\includegraphics[scale=.215]{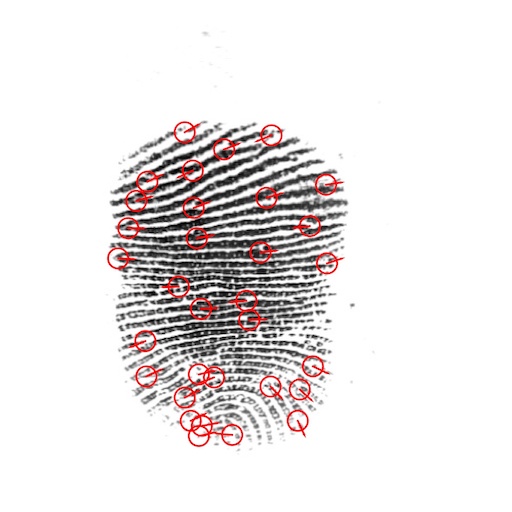}
\end{minipage}%
\begin{minipage}{.2\textwidth}
\centering
\includegraphics[scale=.215]{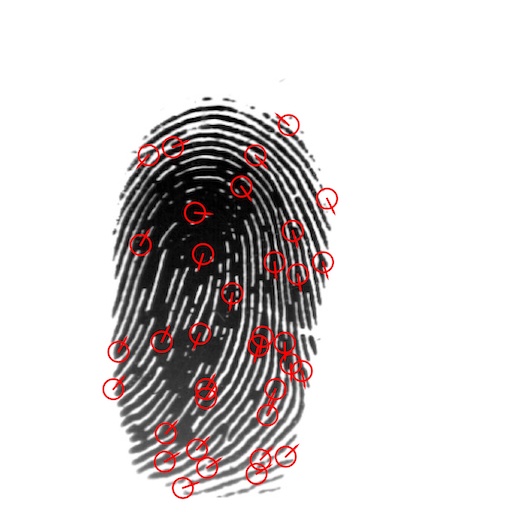}
\end{minipage}%
\centering
\caption{Examples of poor quality fingerprint images from benchmark datasets. Row 1: Rolled fingerprint impressions from NIST SD4. Row 2: Slap fingerprint images from FVC 2004 DB1 A. Rolled fingerprints are often heavily smudged, making them challenging to accurately recognize. FVC 2004 DB1 A also has several distinct challenges such as small overlapping fingerprint area between two fingerprint images, heavy non-linear distortions, and extreme finger conditions (wet or dry). Minutiae annotated with COTS A.}
\label{fig:example_dataset}
\end{figure*}

\subsection{Two-stage DeepPrint Search}

In addition to increasing the speed of large-scale fingerprint search using DeepPrint with product quantization, we also propose a method whereby a negligible amount of search speed can be sacrificed in order to further improve the search accuracy. In particular, we first use the DeepPrint representations to find the top-$k$\footnote{The value of $k$ depends on the gallery size $N$. For the gallery size of $N=1.1$ million, we empirically selected $k=500$.} candidates for a probe $\mathbf{R}_p$ in a gallery $G$. Then, the top-$k$ candidates are re-ranked using the scores of a minutiae-matcher fused together with the DeepPrint similarity scores. More formally, given a minutiae-matcher function $m(.)$, the $k$ re-ranked candidates can be computed by:

\begin{equation}
    Sort_k(\{m(m_p, m_g) + s(\mathbf{R}_p, \mathbf{R}_g) | g \in G\})
    \label{eq:rerank}
\end{equation} where $m_{p}$ and $m_{g}$ two varying length minutiae templates, $\mathbf{R}_{p}$ and $\mathbf{R}_{g}$ are the two fixed-length DeepPrint templates, $s(.)$ is the DeepPrint similarity score (either Equation~\ref{eq:cosine} or Equation~\ref{eq:product_match}), and $Sort_k$ returns a list of $k$ candidates sorted in descending order by similarity score.

We note that since DeepPrint already outputs a minutiae-map, which can easily be converted to a minutiae-set, fusing DeepPrint with a minutiae matcher is quite seamless. We simply convert the DeepPrint minutiae-maps to minutiae-sets, and subsequently input the minutiae-sets to a minutiae-matcher such as the open-source minutiae matcher in~\cite{cao}.

\section{Secure DeepPrint Matching}

One of the primary benefits of the fixed-length, 192-dimensional DeepPrint representation is that it can be encrypted and matched in the encrypted domain (with 192 bits of security~\cite{homomorphic}) with fully homomorphic encryption (FHE). In particular, FHE enables performing any number of both addition and multiplication operations in the encrypted domain. Since DeepPrint representations can be matched using only multiplication and addition operations (Eq.~\ref{eq:cosine}), they can be matched in the encrypted domain with minimal loss to system accuracy (only loss in accuracy comes from converting floating point features to integer value features, resulting in a loss of precision). 

In contrast, minutiae-based representations cannot be matched under FHE, since the matching function cannot be reduced to simple addition and multiplication operations. Furthermore, existing encryption schemes for minutiae-based templates such as the fuzzy-vault, result in a loss of matching accuracy, and are very sensitive to fingerprint pre-alignment~\cite{fuzzy}. We demonstrate in our experiments that the DeepPrint authentication performance remains almost unaltered following FHE matching. We utilize the Fan-Vercauteren FHE Scheme~\cite{homomorphic2} with improvements from~\cite{homomorphic} for improved speed and efficiency\footnote{We use the following open-source implementation:~\url{https://github.com/human-analysis/secure-face-matching}}. 

\begin{table*}[t]
 \centering
\caption{Benchmarking DeepPrint Search Accuracy against Fixed-Length Representations in the Literature and COTS}
\label{table:benchmark_id}
\begin{threeparttable}
\begin{tabular}{ cccccc}
\toprule
Algorithm\tnote{\textdagger} & \specialcell{Template \\Description} & \specialcell{NIST SD4\tnote{1} \\Rank-1 Accuracy (\%)} & \specialcell{NIST SD14\tnote{2} \\Rank-1 Accuracy (\%)} & \specialcell{Template Size \\ Range (kB)} & \specialcell{Search Time \\ (milliseconds)\tnote{3}}\\
\toprule
 \specialcell{Inception v3 + COTS \cite{index1}} & Fixed-Length & 97.80 & N.A. & 8 & 175\\ \\
\specialcell{Finger Patches \cite{index4}} & Fixed-Length & 99.27 & 99.04 & 1.0 & 16 \\ \\
\specialcell{\textbf{DeepPrint}} & Fixed-Length & 98.70 & 99.22 & \textbf{0.2} & \textbf{11}\\ \\
\specialcell{COTS A} & Minutiae-based\tnote{4} & \textbf{99.55} & \textbf{99.92} & (1.5,23.7) & 72 \\ \\
\specialcell{COTS B} & Minutiae-based\tnote{4} & 92.9 & 92.6 & (0.6,5.3) & 20 \\ \\
\bottomrule
\end{tabular}
\begin{tablenotes}
\item[1] Only 2,000 fingerprints are included in the gallery to enable comparison with previous works.
\item[2] Last 2,700 pairs are used to enable comparison with previous works.
\item[3] Search times for all algorithms benchmarked on NIST SD4 with an Intel Core i9-7900X CPU @ 3.30GHz
\item[4] We use the proprietary COTS templates which are comprised of minutiae together with other proprietary features.
\item[\textdagger] These results primarily show that (i) DeepPrints is competitive with the best fixed-length representation in the literature~\cite{index4} (with a smaller template size) and state-of-the-art COTS, but also (ii) the benchmark dataset performances are saturated due to small gallery sizes. Therefore, in subsequent experiments we compare with state-of-the-art COTS against a background of 1.1 million.
\end{tablenotes}
\end{threeparttable}
\end{table*}

\section{Datasets}

We use four sources of data in our experiments. Our training data is a longitudinal dataset comprised of 455K rolled fingerprint images from 38,291 unique fingers taken from~\cite{longitudinal}. Our testing data is comprised of both large area rolled fingerprint images taken from NIST SD4 and NIST SD14 (similar to the training data) and small area slap fingerprint images from FVC 2004 DB1 A.

\subsection{NIST SD4 \& NIST SD14}

The NIST SD4 and NIST SD14 databases are both comprised of rolled fingerprint images (Fig.~\ref{fig:example_dataset}). Due to the number of challenging fingerprint images contained in both datasets (even for commercial matchers), they continue to be popular benchmark datasets for automated fingerprint recognition algorithms. NIST SD4 is comprised of 2,000 unique fingerprint pairs (total of 4,000 images), evenly distributed across the 5 fingerprint types (arch, left loop, right loop, tented arch, and whorl). NIST SD14 is a much larger dataset comprised of 27,000 unique fingerprint pairs. However, in most papers published on fingerprint search, only the last 2,700 pairs from NIST SD14 are utilized for evaluation. To fairly compare DeepPrint with previous approaches, we also use the last 2,700 pairs of NIST SD14 for evaluation.   

\subsection{FVC 2004 DB1 A}

The FVC 2004 DB1 A dataset is an extremely challenging benchmark dataset (even for commercial matchers) for several reasons: (i) small overlapping fingerprint area between fingerprint images from the same subject, (ii) heavy non-linear distortion, and (iii) extremely wet and dry fingers (Fig.~\ref{fig:example_dataset}). Another major motivation for selecting FVC 2004 DB1 A as a benchmark dataset is that it is comprised of slap fingerprint images. Because of this, we are able to demonstrate that even though DeepPrint was trained on rolled fingerprint images similar to NIST SD4 and NIST SD14, our incorporation of domain knowledge into the network architecture enables it to generalize well to slap fingerprint datasets.

\section{COTS Matchers}

In most all of our experiments, we benchmark DeepPrint against COTS A and COTS B (Verifinger 10.0 or Innovatrics v7.2.1.40, the latest version of the SDK as of July, 2019). Due to our Non-disclosure agreement, we cannot provide a link between aliases COTS A and COTS B and Verifinger or Innovatrics. Both of these SDKs provide an ISO minutia-only template as well as a proprietary template comprised of minutiae and other features. To obtain the best performance from each SDK, we extracted the more discriminative proprietary templates. The proprietary templates are comprised of minutiae and other features unknown to us. We note that both Verifinger and Innovatrics are top performers in the NIST and FVC evaluations~\cite{nist, fvc}. 

\section{Benchmark Evaluations}

We begin our experiments by comparing the DeepPrint search performance to the state-of-the-art fixed-length representations reported in the literature. Then, we show that the DeepPrint representation can also be used for state-of-the-art authentication by benchmarking against two of the top COTS fingerprint matchers in the market. We further show that this authentication can be performed in the encrypted domain using fully homomorphic encryption. Finally, we conclude our experiments by benchmarking the large-scale search accuracy of the DeepPrint representation against the same two COTS search algorithms.

\subsection{Search (1:N Comparison)}

Our first experimental objective is to demonstrate that the fixed-length DeepPrint representation can compete with the best fixed-length representations reported in the academic literature~\cite{index1,index4} in terms of its search accuracy on popular benchmark datasets and protocols. In particular, we compute the Rank-1 search accuracy of the DeepPrint representation on both NIST SD4 and the last 2,700 pairs of NIST SD14 to follow the protocol of the earlier studies. 

The results, reported in Table~\ref{table:benchmark_id}, indicate that the DeepPrint representation is competitive with the most accurate search algorithm previously published in~\cite{index4} (slightly lower performance on NIST4 and slightly higher on NIST14). However, we also note that the existing benchmarks (NIST SD4 and NISTSD14) for fingerprint search have now become saturated, making it difficult to showcase the differences between published approaches. Therefore, in subsequent experiments, we better demonstrate the efficacy of the DeepPrint representation by evaluating against a background of 1.1 million fingerprints (instead of the $\approx2K$ in existing benchmarks). 

We highlight once again that DeepPrint has the smallest template among state-of-the-art fixed length representations (200 bytes vs 1,024 bytes for the next smallest).

The search performance on FVC 2004 DB1 A is not reported, since the background is not of sufficient size (only 700 slap prints) to provide any meaningful search results. 

\subsection{Authentication}

We benchmark the authentication performance of DeepPrint against two state-of-the-art COTS minutiae-based matchers, namely COTS A and COTS B. We note that none of the more recent works on fixed-length fingerprint representation~\cite{index1, index2, index3, index4} have considered authentication performance, making it difficult for us to compare with these approaches (to the best of our knowledge, the code for these methods is not open-sourced).

From the experimental results (Tables~\ref{table:auth_fusion} and~\ref{table:benchmark_auth}), we note that DeepPrint outperforms COTS B on all benchmark testing protocols. We further note that DeepPrint outperforms both COTS A and COTS B on the very challenging FVC 2004 DB1 A (Fig.~\ref{fig:example_dataset}). The ability of DeepPrint to surpass COTS A and COTS B on the FVC slap fingerprint dataset is a very exciting find, given the DeepPrint network was trained on rolled fingerprint images which are comprised of very different textural characteristics than slap fingerprint impressions (Fig.~\ref{fig:example_dataset}). In comparison to rolled fingerprints, slap fingerprints often (i) require more severe alignment, (ii) can contain heavier non-linear distortion, (iii) and are much smaller with respect to impression area. We posit that our injection of domain knowledge (both alignment and minutiae detection) into the DeepPrint architecture help it to generalize well from the rolled fingerprints it was trained on to the slap fingerprints comprising FVC 2004 DB1 A. We demonstrate this further in a later ablation study.

\begin{table}[h]
 \centering
\caption{Authentication Accuracy (FVC 2004 DB1 A)}
\label{table:auth_fusion}
\begin{threeparttable}
\resizebox{\columnwidth}{!}{%
\begin{tabular}{cccccc}
\toprule
\specialcell{DeepPrint} &\specialcell{COTS A} &\specialcell{COTS B} & \specialcell{DeepPrint \\+ \\COTS A\tnote{1}} & \specialcell{DeepPrint \\+ \\ COTS B\tnote{1}} & \specialcell{DeepPrint \\+ \\~\cite{cao}\tnote{1}}\\ \\
\hline
\specialcell{97.5\%\tnote{\textdagger}} &\specialcell{96.75\%} &\specialcell{96.57\%} &\specialcell{\textbf{98.93\%}} & \specialcell{98.46\%} & 97.6\%\\ 
\bottomrule
\end{tabular}
}
\begin{tablenotes}
\item[1] Sum score fusion is used.
\item[\textdagger] TAR @ FAR of 0.1\% is reported since there are only 4,950 imposter pairs in the FVC protocol.
\end{tablenotes}
\end{threeparttable}
\vspace{-1.0em}
\end{table}

\begin{table}[h]
 \centering
\caption{Authentication Accuracy (Rolled-Fingerprints)}
\label{table:benchmark_auth}
\begin{threeparttable}
\begin{tabular}{ccc}
\toprule
Algorithm & \specialcell{NIST SD4 \\TAR @ FAR = 0.01\%} & \specialcell{NIST SD14\\TAR @ FAR = 0.01\%} \\
\hline
 \specialcell{COTS A} & \textbf{99.70} & \textbf{99.89} \\ \\
\specialcell{COTS B} & 97.80 & 97.85 \\ \\
\specialcell{Cao \textit{et al.}~\cite{cao}\tnote{1}} & 96.75 & 95.96 \\ \\
\specialcell{\textbf{DeepPrint}} & 97.90 & 98.55 \\ \\
\specialcell{\textbf{DeepPrint} + \\Minutiae\cite{cao}} & 98.70 & 99.0 \\
\bottomrule
\end{tabular}
\begin{tablenotes}
\item[1] Minutiae extracted from DeepPrint minutiae-map ($\mathbf{H}$) and fed directly into minutiae matcher proposed in~\cite{cao}.
\end{tablenotes}
\end{threeparttable}
\vspace{-1.0em}
\end{table}

\subsubsection{Fusion with Minutiae-Matchers}

Another interesting result with respect to the DeepPrint authentication performance is that of the score distributions. In particular, we found that minutiae-based matchers COTS A and COTS B have very peaked imposter distributions near 0. Indeed, this is very typical of minutiae-matchers. In contrast, DeepPrint, has a peaked genuine distribution around 1.0, and a much flatter imposter distribution. In other words, COTS is generally stronger at true rejects, while DeepPrint is stronger at true accepts. This complementary phenomena motivated us to fuse DeepPrint with minutiae-based matchers to further improve their authentication performance (Table~\ref{table:auth_fusion}). Indeed, our results (Table~\ref{table:auth_fusion}) indicate that the DeepPrint representation does contain features complementary to minutiae-based matchers, given the improvement in authentication performance under score level fusion. We note that since a DeepPrints score can be computed with only 192 multiplications and 191 additions, it requires very little overhead for existing COTS matchers to integrate the DeepPrint representation into their matcher.

\subsubsection{Secure Authentication}

In addition to being competitive in authentication accuracy with state-of-the-art minutiae matchers, the \textit{fixed-length} DeepPrint representation also offers the distinct advantage of matching in the encrypted domain (using FHE). Here we verify that the DeepPrint authentication accuracy remains intact following encryption. We also benchmark the authentication speed in the encrypted domain. Our empirical results (Table~\ref{table:auth_encryption}) demonstrate that the authentication accuracy remains nearly the same following FHE, and that authentication between a pair of templates takes only \textbf{1.26} milliseconds in the encrypted domain.

\begin{table}[h]
 \centering
\caption{Encrypted\tnote{1} Authentication using DeepPrint Representation}
\label{table:auth_encryption}
\begin{threeparttable}
\begin{tabular}{cccc}
\toprule
Algorithm & NIST SD4\tnote{2} & NIST SD14\tnote{2} & FVC 2004 DB1 A\tnote{3} \\ \\
\hline
DeepPrint & \specialcell{97.9\%} &\specialcell{98.55\%} &\specialcell{97.5\%} \\ \\

\specialcell{DeepPrint\\+ FHE\tnote{1} }& \specialcell{96.9\%} &\specialcell{97.3\%} &\specialcell{97.0\%} \\
\bottomrule
\end{tabular}
\begin{tablenotes}
\item[1] Fully homomorphic encryption is utilized (match time: 1.26 ms).
\item[2] TAR @ FAR = 0.01\%.
\item[3] TAR @ FAR = 0.1\%
\end{tablenotes}
\end{threeparttable}
\end{table}

\section{Large Scale Search}

Perhaps the most important attribute of the compact DeepPrint representation is its ability to perform extremely fast fingerprint search against large galleries. To adequately showcase this feature, we benchmark the DeepPrint search accuracy against COTS A and COTS B on a gallery of over 1.1 million rolled fingerprint images. The experimental results show that DeepPrint is able to obtain competitive search accuracy with the top COTS algorithm, at orders of magnitude faster speeds. Note, we are unable to benchmark other recent fixed-length representations in the literature against the large scale background, since code for these algorithms has not been open-sourced.

\subsection{DeepPrint Search}

\begin{figure}[h]
\begin{center}
\includegraphics[scale=0.4]{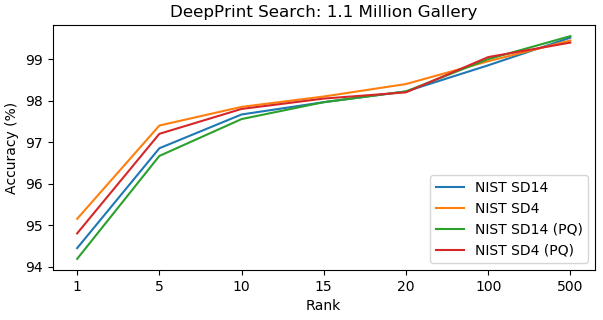}
\caption{Closed-Set Identification Accuracy of DeepPrint (with and without Product Quantization (PQ)) on NIST SD4 and NIST SD14 (last 2,700 pairs) supplemented with a gallery of 1.1 Million. Rank-1 Identification accuracies are 95.15\% and 94.44\%, respectively. Search time is only 160 milliseconds. After adding product quantization, the search time is reduced to 51 milliseconds and the Rank-1 accuracies only drop to 94.8\% and 94.2\%, respectively. }
\label{fig:sd4_deep}
\end{center}
\end{figure}

First, we show the search performance of DeepPrints using a simple exhaustive search technique previously described. In particular, we match a probe template to every template in the gallery, and select the $k$ candidates with the highest similarity scores. We use the NIST SD4 and NIST SD14 databases in conjunction with a gallery of 1.1 million rolled fingerprints. Under this exhaustive search scheme, the DeepPrint representation enables obtaining Rank-1 identification accuracies of \textbf{95.15\% and 94.44\%}, respectively (Table~\ref{table:search_pq}) and (Fig.~\ref{fig:sd4_deep}). Notably, the search time is only \textbf{160 milliseconds}. At Rank-100, the search accuracies for both datasets cross over 99\%. In our subsequent experiments, we demonstrate how we can re-rank the top $k$ candidates to further improve the Rank-1 accuracy with minimal cost to the search time.  

\begin{table}[h]
 \centering
\caption{DeepPrint + Minutiae Re-ranking Search Accuracy (1.1 million background)}
\label{table:search_rerank}
\begin{threeparttable}
\begin{tabular}{c|ccc}
\toprule
Metric & \specialcell{NIST SD4 \\ Rank-1 \\Search Accuracy} & \specialcell{NIST SD14 \\ Rank-1 \\Search Accuracy} & \specialcell{Search Time \\ (milliseconds)\tnote{1}} \\ \\
\hline
\specialcell{DeepPrint \\ +\cite{cao}} & \specialcell{98.8\%} &\specialcell{98.22\%} &\specialcell{\textbf{300}} \\ \\

\specialcell{DeepPrint \\+ COTS A\tnote{2}} & \specialcell{\textbf{99.45}\%} &\specialcell{99.48\%} &\specialcell{11,000} \\ \\

\specialcell{DeepPrint \\+ COTS B\tnote{2}} & \specialcell{98.25\%} &\specialcell{98.41\%} &\specialcell{13,000} \\ \\

\specialcell{COTS A\tnote{3}} & \specialcell{98.85\%} &\specialcell{\textbf{99.51}\%} &\specialcell{27,472} \\ \\

\specialcell{COTS B\tnote{3}} & \specialcell{89.2\%} &\specialcell{85.6\%} &\specialcell{428} \\ 
\bottomrule
\end{tabular}
\begin{tablenotes}
\item[1] Search times benchmarked on an Intel Core i9-7900X CPU @ 3.30GHz
\item[2] COTS only used for re-ranking the top 500 DeepPrint candidates.
\item[3] COTS used to perform search against the entire 1.1 million gallery.
\end{tablenotes}

\end{threeparttable}
\end{table}

\begin{table}[h]
 \centering
\caption{DeepPrint + PQ: Search Accuracy (1.1 million background)}
\label{table:search_pq}
\begin{threeparttable}
\begin{tabular}{cccc}
\toprule
Algorithm & \specialcell{NIST SD4\\Rank 1 \\ Search Accuracy} & \specialcell{NIST SD14\\Rank1\\Search Accuracy} & \specialcell{Search Time \\ (milliseconds)\tnote{1}} \\ \\
\hline
\specialcell{DeepPrint} & \specialcell{95.15\%} &\specialcell{94.44\%} & 160\\ \\

\specialcell{DeepPrint + PQ} & \specialcell{94.80\%} &\specialcell{94.18\%} & \textbf{51}\\ 
\bottomrule
\end{tabular}
\begin{tablenotes}
\item[1] Search times benchmarked on an Intel Core i9-7900X CPU @ 3.30GHz
\end{tablenotes}

\end{threeparttable}
\end{table}

\begin{table}[t]
 \centering
\caption{DeepPrint Representation Comparison}
\label{table:ablation2}
\begin{threeparttable}
\begin{tabular}{c|ccc}
\toprule
Metric & \specialcell{Minutiae \\Representation\tnote{1}} & \specialcell{Texture \\Representation\tnote{1}} & \specialcell{Fused \\Representation\tnote{2}} \\ \\
\hline
\specialcell{FVC 2004 DB1A \\TAR @ \\ FAR = 0.1\%} & \specialcell{97.4\%} &\specialcell{90.0\%} & \specialcell{\textbf{97.5}\%} \\ \\

\specialcell{NIST SD4 \\TAR @ \\ FAR = 0.01\%} & \specialcell{97.0\%} &\specialcell{97.15\%} & \specialcell{\textbf{97.9}\%} \\ \\

\specialcell{NIST SD14 \\TAR @ \\ FAR = 0.01\%} & \specialcell{97.29\%} &\specialcell{98.14\%} & \specialcell{\textbf{98.55}\%} \\
\bottomrule
\end{tabular}
\begin{tablenotes}
\item[1] Each representation (96 bytes) is extracted from one branch in the DeepPrint architecture.
\item[2] Scores from the minutiae representation are fused with the texture representation using sum score fusion.
\end{tablenotes}

\end{threeparttable}
\end{table}

\begin{table}[h!]
 \centering
\caption{DeepPrint Ablation Study}
\label{table:ablation}
\begin{threeparttable}
\begin{tabular}{c|ccc}
\toprule
Metric & w/o all & with alignment & \specialcell{with alignment \\+ domain knowledge} \\ \\
\hline
\specialcell{FVC 2004 DB1A \\TAR @ \\ FAR = 0.1\%} & \specialcell{72.86\%} &\specialcell{88.0\%} &\specialcell{\textbf{97.5\%}} \\ \\

\specialcell{NIST SD4 \\TAR @ \\ FAR = 0.1\%} & \specialcell{96.95\%} &\specialcell{96.65\%} &\specialcell{\textbf{97.9\%}} \\ \\

\specialcell{NIST SD14 \\TAR @ \\ FAR = 0.1\%} & \specialcell{97.96\%} &\specialcell{96.52\%} &\specialcell{\textbf{98.55\%}} \\
\bottomrule
\end{tabular}

\end{threeparttable}
\end{table}

\subsection{Minutiae Re-ranking}

Using the open-source minutiae matcher proposed in~\cite{cao}, COTS A and COTS B, we re-rank the top-$500$ candidates retrieved by the DeepPrint representation to further improve the Rank-1 identification accuracy. Following this re-ranking, we obtain competitive search accuracy as the top COTS SDK, but at significantly faster speeds (Table~\ref{table:search_rerank}).

\subsection{Product Quantization}

We further improve the already fast search speed enabled by the DeepPrint representation by performing product quantization on the templates stored in the gallery. This reduces the DeepPrint template size to only \textbf{64 bytes} and reduces the search speed down to \textbf{51} milliseconds from 160 milliseconds with only marginal loss to search accuracy (Table~\ref{table:search_pq}) and (Fig.~\ref{fig:sd4_deep}).

\begin{figure*}[h]
\begin{center}
\includegraphics[scale=0.24]{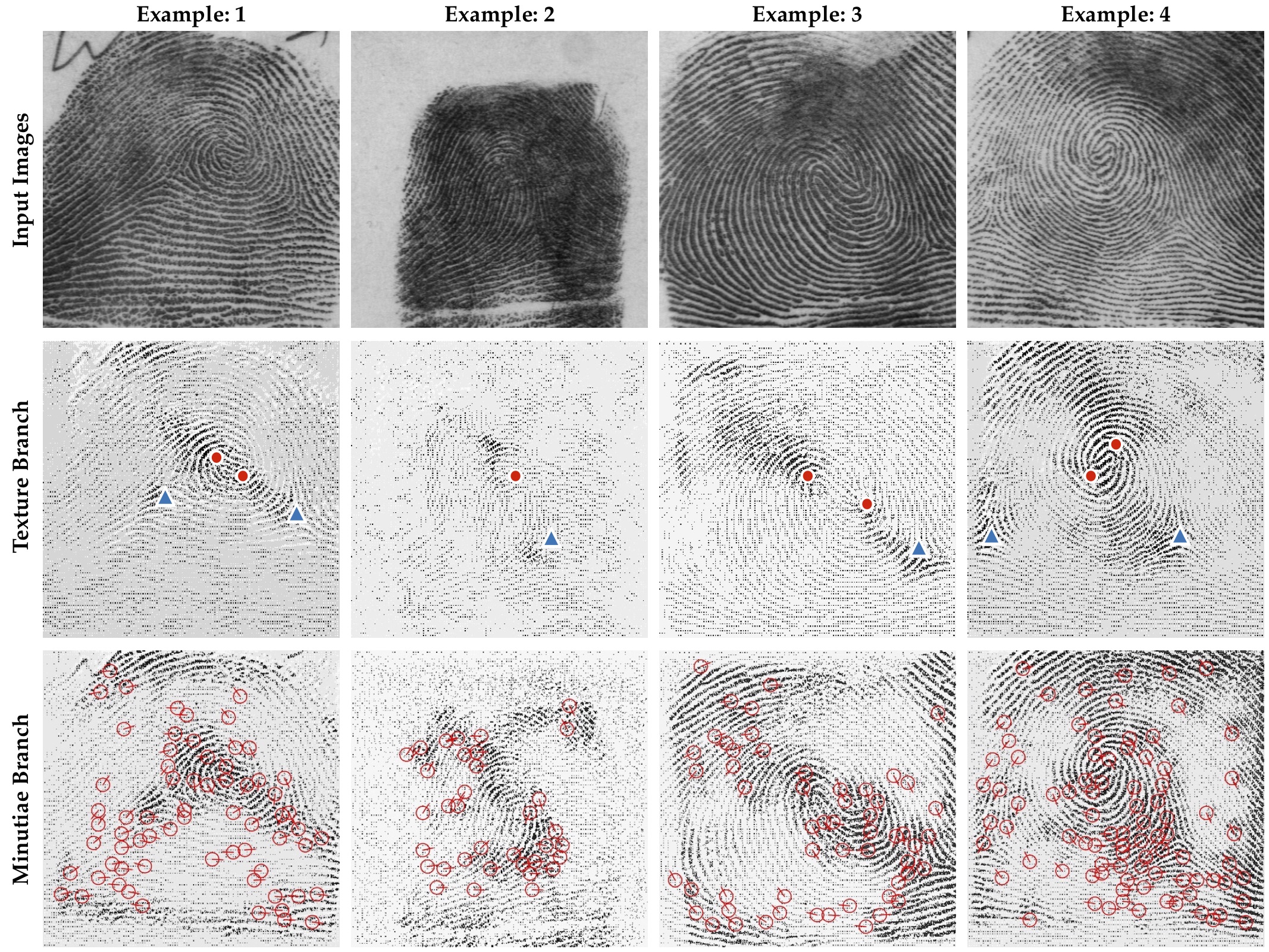}
\caption{Illustration of DeepPrint interpretability. The first row shows three example fingerprints from NIST SD4 which act as inputs to DeepPrint. The second row shows which pixels the texture branch is focusing on as it extracts its feature representation. Singularity points are overlaid to show that the texture branch fixates primarily on regions surrounding the singularity points. The last row shows pixels which the minutiae branch focuses on as it extracts its feature representation. We overlay minutiae to show how the minutiae branch focuses primarily on regions surrounding minutiae points. Thus, each branch of DeepPrint extracts complementary features which comprise more accurate and interpretable fixed-length fingerprint representations than previously reported in the literature.}
\label{fig:interp}
\end{center}
\vspace{-1.0em}
\end{figure*}

\section{Ablation Study}

Finally, we perform an ablation study to highlight the importance of (i) the automatic alignment module in the DeepPrint architecture and (ii) the minutiae-map domain knowledge added during training of the network. In our ablation study, we report the authentication performance of DeepPrint with/without the constituent modules. 

We note that in all scenarios, the addition of domain knowledge improves authentication performance (Tables~\ref{table:ablation2} and \ref{table:ablation}). This is especially true for the FVC 2004 DB1 A database which is comprised of slap fingerprints with different characteristics (size, distortion, conditions) than the rolled fingerprints used for training DeepPrint. Thus we show how adding minutiae domain knowledge enables better generalization of DeepPrint to datasets which are very disparate from its training dataset. We note that alignment does not help in the case of NIST SD4 and NIST SD14 (since rolled fingerprints are already mostly aligned), however, it significantly improves the performance on FVC 2004 DB1 A where fingerprint images are likely to be severely unaligned.

We also note that the minutiae-based representation and the texture-based representation from DeepPrint are indeed complementary, evidenced by the improvement in accuracy when fusing the scores from both representations. (Table~\ref{table:ablation2}).

\section{Interpretability}

As a final experiment, we demonstrate the interpretability of the DeepPrint representation using the deconvolutional network proposed in~\cite{visualizing}. In particular, we show in Fig.~\ref{fig:interp} which pixels in an input fingerprint image are fixated upon by the DeepPrint network as it extracts a representation. From this figure, we make some interesting observations. In particular, we note that while the texture branch of the DeepPrint network seems to only focus on texture surrounding singularity points in the fingerprint (core points, deltas), the minutiae branch focuses on a larger portion of the fingerprint in areas where the density of minutiae points are high. This indicates to us that our guiding the DeepPrint network with minutiae domain knowledge does indeed draw the attention of the network to minutiae points. Since both branches focus on complementary areas and features, the fusion of the representations improves the overall matching performance (Table~\ref{table:ablation2}). 

\section{Computational Resources}

DeepPrint models and training code are implemented in Tensorflow 1.14.0. All models were trained across 2 NVIDIA GeForce RTX 2080 Ti GPUs. All search and authentication experiments were performed with an Intel Core i9-7900X CPU @ 3.30GHz and 32 GB of RAM. 

\section{Conclusion}

We have presented the design of a custom deep network architecture, called DeepPrint, capable of extracting highly discriminative fixed-length fingerprint representations (200 bytes) for both authentication (1:1 fingerprint comparison) and search (1:N fingerprint comparison). We showed how alignment and fingerprint domain knowledge could be added to the DeepPrint network architecture to significantly improve the discriminative power of its representations. Then, we benchmarked DeepPrint against two state-of-the-art COTS matchers on a gallery of 1.1 million fingerprints, and showed competitive search accuracy at significantly faster speeds (300 ms vs. 27,000 ms against a gallery of 1.1 million). We also showed how the DeepPrint representation could be used for matching in the encrypted domain via fully homomorphic encryption. We posit that the compact, fixed-length DeepPrint representation will significantly aid in large-scale fingerprint search. Among the three most popular biometric traits (face, fingerprint, and iris), fingerprint is the only modality for which no state-of-the-art fixed-length representation is available. This work aims to fill this void.

\ifCLASSOPTIONcaptionsoff
  \newpage
\fi



%
\bibliography{ms}
\bibliographystyle{ieeetr}

%

\vspace{-10.0 mm}
\begin{IEEEbiography}[{\includegraphics[width=1in,height=1.25in,clip,keepaspectratio]{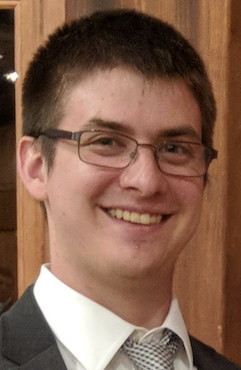}}]{Joshua J. Engelsma}
graduated magna cum laude with a B.S. degree
in computer science from Grand Valley State University, 
Allendale, Michigan, in 2016. He is currently
working towards a PhD degree in the
Department of Computer Science and Engineering
at Michigan State University. His research interests include pattern
recognition, computer vision, and image processing
with applications in biometrics. He won the best paper award at the 2019 IEEE International Conference on Biometrics (ICB).
\end{IEEEbiography}

\vspace{-6.0 mm}
\begin{IEEEbiography}[{\includegraphics[width=1in,height=1.25in,clip,keepaspectratio]{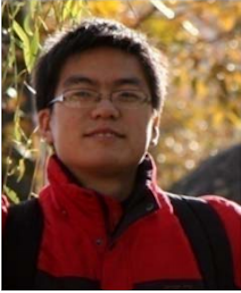}}]{Kai Cao}
received the Ph.D. degree from the
Key Laboratory of Complex Systems and Intelligence
Science, Institute of Automation, Chinese
Academy of Sciences, Beijing, China, in 2010.
He was a Post Doctoral Fellow in the Department
of Computer Science \& Engineering,
Michigan State University. His
research interests include biometric recognition,
image processing and machine learning.
\end{IEEEbiography}

\vspace{-6.0 mm}
\begin{IEEEbiography}[{\includegraphics[width=1in,height=1.25in,clip,keepaspectratio]{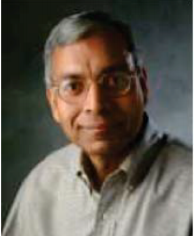}}]{Anil K. Jain} is a University Distinguished Professor in the Department of Computer Science at Michigan State University. He is a Fellow of the ACM, IEEE, IAPR, AAAS and SPIE. His research interests include pattern recognition and biometric authentication. He served as the editor-in-chief of the IEEE Transactions on Pattern Analysis and Machine Intelligence, a member of the United States Defense Science Board and the Forensics Science Standards Board. He has received Fulbright, Guggenheim,  Alexander von Humboldt, and IAPR King Sun Fu awards. He is a member of the United States National Academy of Engineering, a member of The World Academy of Science, and foreign members of the Indian National Academy of Engineering and the Chinese Academy of Sciences.
\end{IEEEbiography}
\vfill




\end{document}